\documentclass[12pt]{article}

\usepackage{latex-common/macros}
\usepackage{paper-macros}
\usepackage[margin=1in]{geometry}

\title{\LARGE \bfseries Which Algorithms \\ Have Tight Generalization Bounds?}

\author{%
	{\normalsize Michael Gastpar} \\
	{\small \textit{EPFL}} \\
	{\small$\mathtt{michael.gastpar@epfl.ch}$}
	\and
	{\normalsize Ido Nachum} \\
	{\small \textit{University of Haifa}} \\
	{\small$\mathtt{inachum@univ.haifa.ac.il}$}
	\and
	{\normalsize Jonathan Shafer} \\
	{\small \textit{MIT}} \\
	{\small$\mathtt{shaferjo@mit.edu}$}
	\vspace*{0.5em}
	\and
	{\normalsize Thomas Weinberger} \\
	{\small \textit{EPFL}} \\
	{\small$\mathtt{thomas.weinberger@epfl.ch}$}
	\vspace*{0.5em}
}

\date{October 2024}

\begin{document}

\pagenumbering{roman}

\maketitle

\thispagestyle{empty}

\begin{abstract}
	We study which machine learning algorithms have tight generalization bounds. First, we present conditions that preclude the existence of tight generalization bounds. Specifically, we show that algorithms that have certain inductive biases that cause them to be unstable do not admit tight generalization bounds. Next, we show that algorithms that are sufficiently stable do have tight generalization bounds.  We conclude with a simple characterization that relates the existence of tight generalization bounds to the conditional variance of the algorithm's loss.  
\end{abstract}

\pagebreak

\thispagestyle{empty}

\tableofcontents

\vfill

\pagebreak

%\input{parts/todos}

%\clearpage

\pagenumbering{arabic}

\section{Introduction}

Generalization bounds are at the heart of learning theory, and they play a central role in attempts to mathematically explain the behavior of contemporary supervised machine learning systems. A generalization bound is an upper bound of the form 
\begin{equation}
    \label{eq:generalization-bound}
    \Loss{\cD}{A(S)} \leq b,
\end{equation}
where $A(S)$ is the hypothesis output by learning algorithm $A$ when executed with training set $S$, and $\Loss{\cD}{\cdot}$ represents the loss with respect to the population distribution $\cD$. The term $b$ is typically an expression of the form 
\begin{equation}
    \label{eq:generalization-bound-simple}
    b = \Loss{S}{A(S)} + c(S, A(S), \cH),
\end{equation}
where $\Loss{S}{\cdot}$ is the empirical loss, $\cH$ is a hypothesis class, and $c(S, A(S), \cH)$ is a `complexity' term, such as the VC dimension or a spectral norm, etc. 

We say that a generalization bound is \emph{valid} if for every population distribution $\cD$, \cref{eq:generalization-bound} holds with high probability; we say that a valid bound is \emph{uniformly tight} (\cref{definition:uniform-tight-bound}) if for every population distribution, with high probability the difference between the two sides of \cref{eq:generalization-bound} is small. 

Bounding the loss using a generalization bound is quite different from using a validation set. Technically, a generalization bound does not use additional samples beyond the training set $S$. And while a validation set provides a single post-hoc measurement of the population loss after training is complete, a good generalization bound can provide insight into \emph{why} a learning algorithm performs well, and can offer guidance for model selection and the development of new learning algorithms. For a generalization bound to be useful in this way, it is important that the bound be tight, so that it can distinguish cases with small population loss from cases with larger loss. 

Unfortunately, experimental works have shown that many of the generalization bounds of the form of \cref{eq:generalization-bound-simple} that have been proposed in the literature are vacuous\footnote{
    A bound is \emph{vacuous} if it is of the form $\PP{\Loss{\cD}{A(S)} \leq b} \geq 1-\delta$ where $\delta \geq 1$ or (for the $0$-$1$ loss) $b \geq 1$. Namely, it is a true statement that provides no guarantees on the performance of the algorithm.  
}
when applied to contemporary learning algorithms such as deep neural networks \parencites{DBLP:conf/iclr/JiangNMKB20}{DBLP:conf/nips/DziugaiteDNRCWM20}[][Section 4.4]{DBLP:conf/aistats/ViallardEHMZ24}.

\textcite{gastpar2023fantastic} offered a partial theoretical explanation for this empirical finding. They considered generalization bound as in \cref{eq:generalization-bound-simple}, namely, bounds that depend only on the training set, the selected hypothesis, and the hypothesis class. They proved that any such bound cannot be uniformly tight in the overparameterized setting.\footnote{They actually showed a stronger result, that such bounds are not tight in an average-case sense for many (algorithms, distribution) pairs.} 
Therefore, they recommended focusing on generalization bounds involving expressions of the form $c(S, A(S), \cH, A, \bbD)$, i.e., bounds that depend also on the specific training algorithm and a specific collection $\bbD$ of population distributions for which the bound is intended.

This recommendation raises the following natural question:
\begin{ShadedBox}
    \begin{question}
        For which algorithms and distribution collections do there exist tight generalization bounds?
    \end{question}
\end{ShadedBox}
In this paper we present conditions that are necessary, sufficient, or necessary and sufficient for the existence of tight generalization bounds for a given learning algorithm and distribution collection.

\subsection{Setting}

Following \textcite{gastpar2023fantastic}, we study the existence of tight generalization bounds using a notion of \emph{estimability}. 

\begin{definition}[Estimability]
    \label{definition:average-estimability}
    Let $\cX$ and $\cY$ be sets, let $m \in \bbN$, let \[
        A: ~ \left(\cX \times \cY\right)^m \to \cY^\cX
    \]
    be a learning rule, and let $\bbD \subseteq \distribution{\cX \times \cY}$ be a collection of distributions. An \ul{estimator} is a function 
    \[
        \estimator: ~ \left(\cX \times \cY\right)^m \to \bbR.
    \]
    Let $\varepsilon, \delta \in [0,1]$. We say that $A$ is \ul{uniformly estimable} (or \ul{worst-case estimable}) with respect to distributions $\bbD$ with precision $\varepsilon$ and confidence $\delta$ using $m$ samples if there exists an estimator $\estimator$ such that
    \begin{equation*}
		\forall \cD \in \bbD: ~ \PPP{
			S \sim \cD^m
		}{
			\bigl|\estimator(S) - \Loss{\cD}{A(S)}\bigr| \leq \varepsilon
		} \geq 1-\delta.
	\end{equation*}
    We say that $A$ is \ul{estimable on average} with respect to distributions $\bbD$ with precision $\varepsilon$ and confidence $\delta$ using $m$ samples if there exists an estimator $\estimator$ such that
    \begin{equation*}
		\PPP{
                    \cD \sim \uniform{\bbD},
                    S \sim \cD^m
		}{
			\bigl|\estimator(S) - \Loss{\cD}{A(S)}\bigr| \leq \varepsilon
		} \geq 1-\delta.
	\end{equation*}
    (More briefly, we say that $(A, \bbD)$ is $(\varepsilon, \delta, m)$-uniformly estimable, or $(\varepsilon, \delta, m)$-estimable on average.)
\end{definition}
The connection between estimability and tight generalization bounds is as follows. 

\begin{fact}
    Using the notation of \cref{definition:average-estimability}, if $(A,\bbD)$ is $(\varepsilon,\delta,m)$-estimable on average, then there exists a generalization bound $b(S)$ (that may depend on $A$ and $\bbD$) that is $\varepsilon$-tight on average, namely
    \begin{equation}
        \label{eq:average-tight-bound}
        \PPP{
			\cD \sim \uniform{\bbD},
                S \sim \cD^m
		}{
			b(S)-\varepsilon \leq \Loss{\cD}{A(S)}\leq b(S)
		} \geq 1-\delta.    
    \end{equation}
    Indeed, the generalization bound is simply $b(S) = \estimator(S)+\varepsilon$, where $\estimator$ is the estimator witnessing the estimability of $(A,\bbD)$. %A similar connection holds between uniform estimability and the existence of uniformly-tight generalization bounds.
    
    In the other direction, if $(A,\bbD)$ is not $(\varepsilon,\delta,m)$-estimable on average, then there exists no generalization bound that satisfies \cref{eq:average-tight-bound}, and in particular no generalization bound can be uniformly tight (as in \cref{definition:uniform-tight-bound}). 
\end{fact}

The main question studied in this paper is as follows: which general and useful conditions are necessary, sufficient, or necessary and sufficient for a tuple $(A, \bbD)$ to be $(\varepsilon, \delta,m)$-uniformly estimable, or $(\varepsilon, \delta,m)$-estimable on average?

We are particularly interested in addressing these questions in settings where the number of samples is not sufficient to guarantee learning in general (in the sense of the VC theorem for example), because most contemporary machine learning algorithms (such as deep neural networks) are used in such settings. This is captured by the following definition.\footnote{
    This is Definition 2 in \textcite{gastpar2023fantastic}. 
    To understand the motivation for this definition, it might be helpful to consider an analogy with the definition of a continuous function. The most common and basic definition of a continuous function is the epsilon-delta definition (developed by Bolzano, Cauchy, Weierstrass and Jordan in the 1800s). The basic definition very clearly captures the intuitive notion of continuity. Later on, however, that definition was generalized by Hausdorff, who gave a modern topological definition of a continuous function in terms of open sets. The modern definition might appear rather strange at first, and somewhat removed from the basic intuition of what a continuous function is. Nonetheless, it turns out that the topological definition does not just generalize the basic definition, but it does so in a way that is very useful, while remaining true to the intuitive notion of continuity.

    In this spirit, Appendix D of \textcite{gastpar2023fantastic} offers a detailed discussion of the merits of \cref{definition:overparameterization}, and how it generalizes many common definitions of overparameterization in a manner that is useful, and is true to the basic intuitive notion of overparameterization.
}
\begin{definition}[Overparameterized setting]
    \label{definition:overparameterization}
    Let $\cX $ and $\cY$ be sets, let $\bbD \subseteq \distribution{\cX \times \cY}$, let $\alpha,\beta \in [0,1]$ and $m \in \bbN$. We say that $(\bbD,m)$ is \ul{$(\alpha,\beta)$-learnable} if there exists a learning rule $A: ~ \left(\cX \times \cY\right)^m \to \cY^\cX$ such that 
    \[
        \PPP{
                \cD \sim \uniform{\bbD}, S \sim (\cD)^m
		}{
			\Loss{\cD}{A(S)} \leq \alpha
        } \geq 1-\beta.
    \]
    We say that $(\bbD,m)$ is \ul{$(\alpha,\beta)$-overparameterized} if it is not $(\alpha,\beta)$-learnable.
\end{definition}

\subsection{Examples}

We present a few simple examples to showcase the richness of the estimability setting. 
In this section $\varepsilon,\delta \in (0,1)$, $\cX$ is a set, $m \in \bbN$ is a sample size, $A: \: \left(\cX \times \pmo\right)^m \allowbreak \to \pmo^\cX$ is a learning rule, and $S = ((x_1,y_1), \dots, (x_m, y_m))$ is a training set.

\begin{example}[Perfect learnability does not imply perfect estimability]
    Let $\cX = [0,1]$, let $\bbD = \distribution{\cX \times \{1\}}$ be the set of all distributions of labeled examples $(x,y)$ where $x \in \cX$ and $y=1$. The collection $\bbD$ is perfectly learnable, that is, there exists a learning algorithm that always achieves $0$ population loss (namely, the learning algorithm that always outputs the constant function $h(x) = 1$). 

    Nonetheless, not every learning algorithm is worst-case estimable with respect to $\bbD$. Indeed, consider the algorithm $A$ that on input $S$ outputs the hypothesis 
    \[
        h(x) =
        \left\{
        \begin{array}{ll}
        	-1  & x \in \{x_1,\dots,x_m\} \\
        	+1 & \text{otherwise}.
        \end{array}
        \right.
    \]
    For any distribution $\cD \in \bbD$, $\Loss{\cD}{A(S)} = \cD_{\cX}(\{x_1,\dots,x_m\})$, where $\cD_{\cX}$ is the marginal of $\cD$ on $\cX$. 
    Hence, estimating the loss of $A$ is equivalent to a task of support size estimation, which is difficult.  Concretely, for any finite set $T \subseteq \cX$, let $\cD_T = \uniform{T \times \{1\}}$. Let $\estimator$ be any estimator, and consider an experiment where with probability $1/2$, we sample $T \sim \uniform{\cX}^{m^2}$ and set $\cD = \cD_T$, and with probability $1/2$ we set $\cD = \cD_{\mathrm{U}} := \uniform{\cX \times \{1\}}$. Consider the probability 
    \[
        p = \PPP{
			S \sim \cD^m
		}{
			\Bigl|\estimator(S) - \Loss{\cD}{A(S)}\Bigr| \geq \frac{1}{2m}
        }.
    \]
    Let $E$ be the event where $|\{x_1,\dots,x_m\}|=m$. In the case where $\cD = \cD_T$ with $|T| = m^2$, \cref{claim:birthday-paradox} implies that $\PP{E} \geq 1/e$. And in the case where $\cD = \cD_{\mathrm{U}}$, $\PP{E} = 1$. Hence, in both cases, with probability at least $1/e$, the estimator receives a sample of $m$ distinct points chosen independently and uniformly from $\cX$, and it cannot distinguish between these two cases. However, $\Loss{\cD_{\mathrm{U}}}{A(S)} = 0$, whereas $\Loss{\cD_T}{A(S)} = \frac{1}{m}$ when $E$ occurs. This implies that $p \geq 1/2e$, and so $(A,\bbD)$ is not $(\frac{1}{2m}, \delta,m)$-uniformly estimable for any $\delta < 1/2e$.\hfill$\square$
\end{example}

Some algorithms are very estimable but are not good learning algorithms, as in the following three examples.

\begin{example}[Constant algorithms are estimable]
    \label{example:constant-algorithm}
    Let $m \geq \log(1/\delta)/\varepsilon^2$. Let $h_0: \: \cX \to \pmo$ be a function, and let $A$ be the constant learning algorithm such that $A(S) = h_0$ for all $S$. Then by Hoeffding's inequality, $A$ is $(\varepsilon,\delta,m)$-uniformly estimable with respect to the set of all distributions $\bbD = \distribution{\cX \times \pmo}$, with estimator $\estimator(S) = \Loss{S}{h_0}$.\hfill$\square$
\end{example}

For some algorithms, the empirical loss is not a good estimator, yet the algorithm is still estimable.

\begin{example}[Memorization]
    \label{example:memorization}
    Let $\OmegaOf{\log(1/\delta)/\varepsilon^2} \leq m \leq \BigO{\varepsilon |\cX|}$, and consider the algorithm $A$ that on input $S$, outputs the hypothesis  
    \[
        h(x) =
        \left\{
        \begin{array}{ll}
        	y  & \exists y: \: \{y\} = \{y_i: \: i \in [m] \: \land \: x_i = x\} \\
        	-1 & \text{otherwise}.
        \end{array}
        \right.
    \]
    Let $\bbD$ be the collection of all distributions over $\cX \times \pmo$ that have a uniform marginal on $\cX$. Note that $A$ always has $0$ empirical loss. However, $(A,\bbD)$ is $(\varepsilon,\delta)$-uniformly estimable, using $\estimator(S) = \left|\{i \in [m]: ~ y_i = 1\}\right|/m$.\hfill$\square$ 
\end{example}

\begin{example}[Most algorithms are estimable] 
    \label{example:random-algorithm}
    Let $d = |\cX| < \infty$, let $\cF = \pmo^\cX$, and for each $f \in \cF$, let $\cD_f = \uniform{\left\{\left(x, f(x)\right): \: x \in \cX\right\}}$.
    Let $\cA$ be the set of all mappings $\left(\cX \times \pmo\right)^m \allowbreak \to \pmo^\cX$, and consider a mapping $A$ chosen uniformly from the set $\cA$. For any fixed $f \in \cF$ and for any fixed sample $S$ of size $m$ consistent with $f$, $A(S)$ is a function that was chosen uniformly from $\cF$. By Hoeffding's inequality,
    \[
        \forall f \in \cF ~ \forall S \in \supp(\cD_f): ~ 
        \PPP{
            A \sim \uniform{\cA}
        }{
            \left|\Loss{\cD_f}{A(S)}-\frac{1}{2}\right| \geq \varepsilon
        } \leq 2e^{-2d\varepsilon^2}.
    \]
    In particular,
    \[
        \PPP{
            A \sim \uniform{\cA}, 
            f \sim \uniform{\cF}, 
            S \sim \left(\cD_f\!\right)^m
        }{
            \left|\Loss{\cD_f}{A(S)}-\frac{1}{2}\right| \geq \varepsilon
        } \leq 2e^{-2d\varepsilon^2}.
    \]
    Hence, by Markov's inequality, 99\% of algorithms $A \in \cA$ satisfy that $(A, \{\cD_f\}_{f \in \cF})$ is $(\varepsilon, 200e^{-2d\varepsilon^2},m)$-estimable on average.\hfill$\square$
\end{example}

A similar argument shows also that most ERM algorithms are estimable in the overparameterized setting.\footnote{
    Consider an overparameterized setting with $m=o(d)$. The output of any (realizable) ERM will have zero error on points in $S$, and will make an error on each unseen data points with probability $1 / 2$, yielding an expected population loss of $(d-m) / 2 d \approx 1 / 2$. Hence, essentially the same result as in \cref{example:random-algorithm} can be obtained also for ERMs by applying Hoeffding's inequality.
} 
In both cases, the algorithms are estimable because their loss is guaranteed to be high, namely, the algorithms are poor learners. 

Finally, algorithms for learning parity functions are a particularly instructive case.

\begin{example}[Parity functions]
    \label{example:parities}
    Let $d \in \bbN$ be large enough, $\cX = (\bbF_2)^d$, and let $\cH = \left\{f_w: ~ w \in \cX\right\} \subseteq (\bbF_2)^\cX$ be the class of parity functions such that $f_w(x) = \sum_{i\in[d]} w_i \cdot x_i$. 
    Let $\bbD = \{\cD_f\}_{f \in \cH}$ with $\cD_f = \uniform{\{(x,f(x)): \: x \in \cX\}}$.
    % Let $V_1 \subseteq V_2 \subseteq \cdots \subseteq V_d = (\bbF_2)^d$ be a sequence of linear subspaces such that for each $i \in [d]$, $\dim(V_i)=i$. Consider an ERM algorithm $A$ for $\cH$ that on input $S$, outputs a function $f_w$ consistent with $S$ such that $w \in V_{k}$ for $k$ minimal.
    For a learning rule $A$ and sample size $m$, let
    \[
        p(m) = \PPP{
			\substack{\cD \sim \uniform{\bbD} 
			\\ 
			S \sim (\cD)^m}
		}{
			\Loss{\cD}{A(S)} = 0
        }.
    \]

   For sample size $m\geq d+10$, any ERM algorithm  for $\cH$ satisfies $p(m) \geq 0.999$, meaning it learns $\bbD$ well, and hence is $(0,10^{-3},d+10)$-estimable on average. This holds because for an ERM to output the ground truth, it is clearly sufficient that only a single sample-consistent function exists in the concept class (the ground truth). Similarly, in the event that there are $t>1$ sample-consistent functions, the success probability is given by $1/t$ due to the uniform prior over ground truth distributions. Parity functions are fully characterized by their coefficient vector $w=[w_1,\dots,w_d]$. Since the labels $y$ are a bilinear function in the inputs $x$ and coefficients $w$, one can obtain $w$ from $m\geq d$ linearly independent samples ${x_i}$ by solving the linear system of equation $y=Xw$ with design matrix $X\in \{0,1\}^{m \times d}$. More generally, $X$ having rank $d-k$ is equivalent to the event of having $t=2^k$ sample-consistent functions (coefficient vectors) since every additional linearly independent row rules out half of all parity functions. Now assume $X$ consists of all i.i.d. Ber(½) entries and $y$ contains the labels of all samples. The probability of zero population loss can now be obtained from the law of total probability with the probabilities of rank deficiency computed according to Corollary 2.2 in \textcite{blake2006properties}.

    Similarly, for smaller sample sizes, any ERM for $\cH$ satisfies $p(d) \geq 0.61$, and $p(d-1) \geq 0.38$.
    However, ERM algorithms differ in their degree of estimability for smaller sample sizes. Concretely, Theorem 5 in \textcite{gastpar2023fantastic} shows that there exist ERM algorithms such that for any $6 \leq m \leq d$ there exists a collection $\bbD_m$ for which the algorithm is not $(0.25, 0.32, m)$-estimable on average. These algorithms have an inductive bias towards a subset $\cF \subseteq \cH$, such that they perform well for distributions labeled by a function from $\cF$, and perform poorly for target functions from the complement of $\cF$. In contrast, for the same hard collections $\bbD_m$, ERM algorithms without an inductive bias perform poorly on all distributions for small $m$, so they are significantly more estimable. \hfill$\square$
\end{example}

ERM algorithms for parity functions demonstrate two important phenomena: (1) Estimability can be a very delicate matter, in the sense that changing the sample size by a small additive constant can make all the difference (e.g., any ERM for parities is very estimable with $m=d+10$ samples, but not very estimable with $m=d$); (2) when the sample size is not sufficient for learning all the distributions in the collection $\bbD$, there can be a trade-off between learning performance and estimability. Algorithms with no inductive bias will perform equally poorly for all distributions, and this makes them estimable. In contrast, algorithms that have an inductive bias towards a subset $\bbD' \subseteq \bbD$ can perform well on $\bbD'$, and this can make them less estimable.

\subsection{Our Results}

We investigate which algorithms and collections of distributions are estimable. 
Recall that in \cref{example:parities} we saw that estimability is a delicate phenomenon. In particular, changing the sample size by just a small constant number can in some cases drastically change the set of $(\varepsilon,\delta)$ estimability parameters that are achievable. This means that identifying a simple and tight characterization that precisely determines the number of samples necessary and sufficient for estimability can be a difficult undertaking. 

In this paper, we present conditions that preclude estimability, conditions that guarantee estimability, and a condition that is both necessary and sufficient for estimability.

Our first result is a condition that precludes estimability for algorithms that have an inductive bias towards certain subsets of VC classes, showing a connection between estimability and a central notion from traditional learning theory.
\begin{theorem*}[Informal version of \cref{theorem:vc-class-estimability-bound}]
    \emph{
    Let $\cH \subseteq \pmo^\cX$ be a hypothesis class with VC dimension $d$ large enough, and let $m \leq \sqrt{d}/10$. Then there exists a subset $\cF \subseteq \cH$ and corresponding realizable distributions $\bbD$ such that any learning rule that has an inductive bias towards $\cF$ is not $(1/4-o(1), 1/6, m)$-estimable on average over $\bbD$.
    }
\end{theorem*}
Note that the theorem precludes estimability on average, and so in particular it precludes worst-case estimability. The proof of \cref{theorem:vc-class-estimability-bound} uses the Johnson--Lindenstrauss lemma (\cref{theorem:johnson-lindenstrauss}), the probabilistic method, and a technical lemma (\cref{lemma:technical}) concerning the estimability of nearly-orthogonal functions. 

Our next inestimability result is as follows.
\begin{theorem*}[Informal version of \cref{theorem:orthogonal-functions}]
    \emph{
    Let $\cH \subseteq \pmo^\cX$ be a collection of roughly $2^m$ nearly-orthogonal functions and corresponding realizable distributions $\bbD$. Then any learning rule that has an inductive bias towards $\cH$ is not $(1/4-o(1), \sim 1/6, m)$-estimable on average over $\bbD$.
    }
\end{theorem*}
\cref{theorem:orthogonal-functions} is partially stronger than \cref{theorem:vc-class-estimability-bound} in the sense that it shows inestimability for \emph{every} algorithm that has an inductive bias towards a class of nearly-orthogonal functions, whereas \cref{theorem:vc-class-estimability-bound} only shows the existence of a subclass with this property.\footnote{
    Additionally the quantity hidden by the $o(1)$ notation is smaller in \cref{theorem:orthogonal-functions} by a quadratic factor (order $1/m$ vs.\ $1/\sqrt{m}$).
}
On the other hand, \cref{theorem:vc-class-estimability-bound} is stronger than \cref{theorem:orthogonal-functions} in the sense that if \cref{theorem:orthogonal-functions} is applied to show inestimability for subclasses of a VC class, then it yields inestimability only for $m \leq \BigO{\sqrt[3]{d}}$, whereas \cref{theorem:vc-class-estimability-bound} obtains inestimability for all $m \leq \BigO{\sqrt{d}}$.\footnote{
    The limitation $m \leq \BigO{\sqrt[3]{d}}$ when using \cref{theorem:orthogonal-functions} follows from the tightness of the Johnson--Lindenstrauss (JL) lemma. By the JL lemma, taking a collection $\cF$ of $2^m$ orthogonal functions on a high dimensional domain, we can project $\cF$ using a random projection and obtain a collection $\cF'$ of $2^m$ functions that are $\varepsilon$-orthogonal defined on a domain of dimension $\log \left(2^m\right) / \varepsilon^2$. In particular, let $\cH$ be a class with VC dimension $d$. We want to project $\cF$ onto an $\cH$-shattered set of size $d$ with $\varepsilon=\Theta(1 / m)$. This yields $d=m /(\Theta(1 / m))^2=\Theta\left(m^3\right)$. The tightness of JL implies that this construction cannot be improved.
}

To show \cref{theorem:orthogonal-functions}, we prove a concentration inequality using the duality of linear programs (\cref{lemma:lp-concentration}), and then invoke the technical lemma (\cref{lemma:technical}).

One way to interpret \cref{theorem:vc-class-estimability-bound,theorem:orthogonal-functions} is to consider a scenario where one derives a new generalization bound for a given algorithm, without making explicit distributional assumptions (as is the case for many published generalization bounds), and having a sample size within the regime of our theorems. Such bounds are generally formulated as high probability upper bounds on the population loss. Note that the lack of distributional assumptions means that the bound has to hold (be a valid upper bound) for all distributions, including the families of distributions that appear in our theorems.

But this means, in the light of our theorems, that the considered bound is necessarily very weak for many distributions unless one satisfies at least one of the following items:

\begin{enumerate}
    \item{
        Exclude in advance all families of distributions with nearly-orthogonal labeling functions, and use this fact in the derivation of the generalization bound.
    }
    \item{
        Mathematically show that the algorithm is not biased towards any set of nearly-orthogonal functions.\footnote{
            It is known that there exist at least some neural network architectures which, when trained with SGD, are capable of learning orthogonal functions (such as parities). See Theorem 1 in \textcite{colin}.
        }
    }
\end{enumerate}

The intuition behind \cref{theorem:orthogonal-functions} is that having an inductive bias towards a collection $\cH$ of nearly-orthogonal functions makes the algorithm very unstable -- small changes in the training set will cause the algorithm to shift between hypotheses in $\cH$, which are all very different from one another. This motivates our next result, which shows that stable algorithms are estimable, as follows.
\begin{theorem*}[Informal version of \cref{theorem:stablity-implies-estimability}]\emph{
        Let $A$ be an algorithm that is sufficiently stable with respect to a collection of distributions $\bbD$ (in a sense of loss stability or hypothesis stability similar to \cite{rogers1978finite}, or \cite{DBLP:journals/neco/KearnsR99}). Then $(A,\bbD)$ is estimable. 
    }
\end{theorem*}

An additional motivation for \cref{theorem:stablity-implies-estimability} is the intuition that contemporary machine learning algorithms (like deep neural networks) might indeed be sufficiently stable. If so, \cref{theorem:stablity-implies-estimability} would apply, meaning that it is possible to obtain tight generalization bounds for deep neural networks based on the stability property. To substantiate this intuition, we conduct simple preliminary experiments to estimate the the stability of neural networks in practice. Our empirical findings, presented in \cref{section:experiments}, suggest that neural networks are indeed quite stable.  

Finally, in \cref{section:characterization}, we present a necessary and sufficient condition for estimability based on the conditional variance of the algorithm's loss. This characterization is formalized in terms of $\ell_2$ estimability, which is asymptotically equivalent to average case estimability via Markov's inequality.  

\begin{fact*}[\cref{fact:tautology}]\emph{
         $A$ is $(\varepsilon,m)$-estimable in $\ell_2$ with respect to $\bbD$  if and only if
        \[
        \EE{ \operatorname{var}(L_{\mathcal{D}}({A}(S)) \mid S)} \leq \varepsilon.
        \] 
    }
\end{fact*}

\subsection{Related Works}

The works of \textcite[][Theorem 3.1]{nagarajan2019uniform} and \textcite[][Theorem 1]{bartlett2021failures} also study cases where generalization bounds fall short of estimating the performance of learning algorithms (while \cite{negrea2020defense} provide a response to these claims). They preclude tight algorithm-dependent generalization bounds only for uniform convergence and linear classifiers. Their theorems consider specific distributions (Gaussian in \cite{nagarajan2019uniform}, a different distribution per sample in \cite{bartlett2021failures}) and specific types of SGD. In contrast, our work uses the same marginal distribution across all sample sizes, and applies to many algorithms and distributions.

Our definition of an estimator $\estimator$ formalizes algorithm-dependent generalization bounds. We mention several results from the literature:
\textcite{zhang2023lower} studies convex optimization, so the results apply only to a single neuron. While providing matching lower and upper bounds, these bounds match only asymptotically when the sample size $n$ is very large, far from the overparameterized regime relevant for neural networks. \textcite{nikolakakis2023beyond} proposes new generalization bounds for algorithms satisfying a certain symmetry property (e.g., full-batch gradient descent) when using smooth losses. These bounds are algorithm-dependent but distribution-free, making no distributional assumptions.

The following are information-theoretic generalization bounds that are both algorithm and distribution-dependent. However, such bounds are sometimes difficult to approximate numerically in a tight manner (e.g., the bound in Theorem 1 of \cite{xu2017information}). These bounds are part of the PAC-Bayes framework; see proof 4 for Theorem 8 in \textcite{bassily2018learners}, which is equivalent to Theorem 1 in \textcite{xu2017information}. Unfortunately, when these PAC-Bayes or information-theoretic bounds can be approximated in a tight manner such as in \textcite{IssaEG:2019}, \textcite{pmlr-v195-issa23a}, \textcite{EspositoGI:2021}, \textcite{harutyunyan2021information}, \textcite{dziugaite2021role}, \textcite{Haghifam2022}, \textcite{hellstrom2022new}, and \textcite{wang2023tighter}, they do not reveal what properties of the (distribution, algorithm) pair allowed for such success in learning and estimation. The works of \textcite{info_stab1} and \textcite{info_stab2} use the notion of leave-one-out conditional mutual information to derive generalization bounds, which provide another characterization of VC classes and yield non-vacuous generalization bounds for neural networks.

In this paper, we formalize a simple stability condition that guarantees the existence of tight generalization bounds. This condition shares a similarity to the definition of hypothesis stability and loss stability in \textcite{DBLP:journals/neco/KearnsR99}, \textcite{stable_elisseeff05a}, and \textcite{rogers1978finite}. \textcite{stab_Lei2022StabilityAG}
use another similar definition for stability and utilize it to derive generalization bounds for GD and SGD.

Our definitions of stability are also reminiscent of the replace-one stability in \textcite{bousquet2002stability}, but as we explain in \cref{sec:stab}, that definition has some limitations.

\subsection{Preliminaries}

\begin{definition}
    For $m \in \bbN$ and sets $\cX$ and $\cY$, a \ul{learning rule} is a function $A: ~ \left(\cX \times \cY\right)^m \to \cY^\cX$. We will also consider learning rules with variable-size input, i.e., $A: ~ \left(\cX \times \cY\right)^* \to \cY^\cX$.
\end{definition}

In this paper we informally use the terms `learning algorithm' and `learning rule' interchangeably. Both words refer to a function, ignoring considerations of computability. All learning algorithms in this paper are deterministic.

\begin{notation}
    For a set $\Omega$, we write $\distribution{\Omega}$ to denote the collection of all probability measures over a measurable space $(\Omega, \cF)$, where $\cF$ is some fixed $\sigma$-algebra that is implicitly understood. We write $\uniform{\Omega}$ to denote the uniform distribution over $\Omega$. 
\end{notation}

\begin{definition}
    Let $m \in \bbN$, let $\cX$, $\cY$ be sets, let $h: \: \cX \to \cY$, let $S = ((x_1,y_1),\dots,(x_m,y_m)) \in (\cX \times \cY)^m$, and let $\cD \in \distribution{\cX \times \cY}$. The \ul{empirical loss of $h$ with respect to $S$} is
    $\Loss{S}{h} = \frac{1}{m}\sum_{i \in [m]} \indicator{h(x_i) \neq y_i}$. The \ul{population loss of $h$ with respect to $\cD$} is $\Loss{\cD}{h} = \PPP{(x,y) \sim \cD}{h(x) \neq y}$.
\end{definition}

\begin{definition}[Uniformly tight generalization bound for an algorithm]
    \label{definition:uniform-tight-bound}
    Let $m \in \bbN$, $\varepsilon, \delta \in [0,1]$, let $\cX$ and $\cY$ be sets, let $m \in \bbN$, let $A: ~ \left(\cX \times \cY\right)^m \to \cY^\cX$
    be a learning rule, and let $b: \left(\cX \times \cY\right)^m \to [0,1]$ be a generalization bound (that may depend on $A$). We say that $b$ is \ul{uniformly tight for $A$ with precision $\varepsilon$ and confidence $\delta$} if for any distribution $\cD \in \distribution{\cX \times \cY}$,
	\begin{equation*}
        \PPP{
			S \sim \cD^m
		}{
			b(S)-\varepsilon \leq \Loss{\cD}{A(S)}\leq b(S)
		} \geq 1-\delta.    
    \end{equation*}
\end{definition}
\begin{notation}
    Let $\cX$ be a set, let $\cF \subseteq \pmo^\cX$ be a hypothesis class, and let $S \in \left(\cX \times \pmo\right)^*$. We denote $\cF_S = \{f \in \cF: ~ \Loss{S}{f} = 0\}$.
\end{notation}

The following definition captures the notion of a learning rule having an inductive bias towards a particular set of hypotheses.

\begin{definition}
	Let $m \in \bbN$, let $\cX$ be a set, and let $\cF \subseteq \pmo^\cX$ be a hypothesis class. We say that a learning rule $A: ~ \left(\cX \times \pmo\right)^m \to \pmo^\cX$ is \ul{$\cF$-interpolating} if $A(S) \in \cF_S$ for every sample $S \in \left(\cX \times \pmo\right)^m$ such that $\cF_S \neq \varnothing$.
\end{definition}

\begin{remark}
	The property of $\cF$-interpolation is similar to the more common property of \emph{proper empirical risk minimization (proper ERM)} for $\cF$. However, $\cF$-interpolation is a slightly weaker requirement. Specifically, if $S$ is not $\cF$-realizable (i.e., $\cF_S = \varnothing$), then an $\cF$-interpolating learning rule may output any function in $\pmo^\cX$, whereas a proper learning rule for $\cF$ must always output a function from $\cF$.
\end{remark}

\begin{definition}
	\label{definition:eta-orthogonal}
	Let $\varepsilon \geq 0$, let $\cX$ be a set, and let $\cF \subseteq \pmo^\cX$ be a hypothesis class. We say that $\cF$ is \ul{$\varepsilon$-orthogonal with respect to $\cX$}, denoted $\cF \in \orthfull{\varepsilon}{\cX}$, if every distinct $f,g \in \cF$ satisfy
	\[
		\left|\EEE{x \sim \uniform{\cX}}{f(x)g(x)}\right| \leq \varepsilon.	
	\]
	For simplicity, we write $\cF \in \orth{\varepsilon}$ when $\cX$ is understood from context. 
\end{definition}

\begin{fact}
	\label{fact:orthogonal-loss}
	Let $\varepsilon > 0$ and let $\cF \subseteq \pmo^\cX$ be $\varepsilon$-orthogonal. Then for any distinct $f,g \in \cF$,
	\begin{align*}
		\frac{1}{2} - \frac{\varepsilon}{2} \leq \PPP{x \sim \uniform{\cX}}{f(x) = g(x)} \leq \frac{1}{2} + \frac{\varepsilon}{2}.
	\end{align*} 
\end{fact}

\begin{proof}\quad
	$\begin{aligned}[t]
		\PPP{x \sim \uniform{\cX}}{f(x) = g(x)} 
		&=
		\EEE{x \sim \uniform{\cX}}{\indicator{f(x) = g(x)}} 
		=
		\EEE{x \sim \uniform{\cX}}{\frac{1 + f(x)g(x)}{2}} 
		\\
		&=
		\frac{1}{2} + \frac{1}{2}\cdot\EEE{x \sim \uniform{\cX}}{f(x)g(x)}. && \qedhere
	\end{aligned}$
\end{proof}

\section{Conditions that Preclude Estimability}

We present two conditions that preclude estimability. 

\subsection{Inestimability for VC Classes}

\begin{theorem}
	\label{theorem:vc-class-estimability-bound}
	There exists $d_0>0$ as follows. For any integer $d \geq d_0$, let $\cX$ be a set, let $\cH \subseteq \pmo^\cX$ such that $\VC{\cH} = d$, and let $m \in \bbN$ such that $m\leq\sqrt{d}/10$.
	Then there exists a subset $\cF \subseteq \cH$ and a collection $\bbD \subseteq \distribution{\cX \times \pmo}$ of $\cF$-realizable distributions such that for any $\cF$-interpolating learning rule $A$ and for any estimator $\estimator: ~ (\cX \times \pmo)^m \to [0,1]$ that may depend on $\bbD$ and $A$, 
	\begin{equation}
		\label{eq:vc-class-not-estimable}
		\PPP{
			\substack{\cD \sim \uniform{\bbD} 
			\\ 
			S \sim \cD^m}
		}{
			\bigl|\estimator(S) - \Loss{\cD}{A(S)}\bigr| \geq \frac{1}{4}-\frac{1}{2d^{1/4}}
		} \geq \frac{1}{6}.
	\end{equation}
\end{theorem}

The proof of \cref{theorem:vc-class-estimability-bound} appears in \cref{appendix:proof-vc}. We note that some of the constants appearing in the theorem were chosen for simplicity, and can be improved.

\subsection{Inestimability for Nearly-Orthogonal Functions}

\begin{theorem}
    \label{theorem:orthogonal-functions}
	Let $m \in \bbN$, let $\cX$ be a set, and let $A: ~ \left(\cX \times \pmo\right)^m \to \pmo^\cX$ be a learning rule.
	Assume that $A$ is $\cF$-interpolating for a set $\cF \subseteq \pmo^{\cX'}$ where $\cX' \subseteq \cX$, $100m^2 \leq |\cX'| < \infty$, $\cF \in \orthfull{1/1000m}{\cX'}$ and $|\cF| = 2^m+1$. 
	Then there exists a collection of $\cF$-realizable distributions $\bbD \subseteq \distribution{\cX' \times \pmo}$ such that for any estimator function $\estimator: ~ (\cX \times \pmo)^m \to [0,1]$ that may depend on $\bbD$ and $A$, 
	\[
		\PPP{
			\substack{\cD \sim \uniform{\bbD} 
			\\ 
			S \sim \cD^m}
		}{
			\bigl|\estimator(S) - \Loss{\cD}{A(S)}\bigr| \geq \frac{1}{4}-\frac{1}{4000m}
		} \geq 0.16.
	\]
\end{theorem}

The proof of \cref{theorem:orthogonal-functions} appears in \cref{appendix:proof-orthogonal}. We note that here too, the constants appearing in the theorem were chosen for simplicity, and can be improved. In particular, using a similar technique it is possible to show a lower bound of $1/6$ instead of $0.16$, matching the bound in \cref{theorem:vc-class-estimability-bound}.

\section{Sufficient Conditions for Estimability}
\label{sec:stab}

In \cref{example:constant-algorithm,example:memorization} we saw that the constant algorithm and the memorization algorithm are very estimable. These algorithms are also very stable. Indeed, they always output the same (or essentially the same) hypothesis.\footnote{The memorization algorithm always outputs the function $h(x)=-1$, except that it alters $h$ in a small number of locations to fit the training set.} In the other direction, \cref{theorem:orthogonal-functions} shows that certain algorithms that are very unstable, are not estimable. This suggests that stability might play an important role in determining the estimability of an algorithm.

One notion of algorithmic stability that is common in the literature is leave-one-out stability \parencite{bousquet2002stability}. However, it is easy to see that the memorization algorithm, which is estimable and is (intuitively) very stable, does not satisfy this definitions of stability. 
Therefore, we use the following alternative definitions of algorithmic stability, which are similar to \textcites{rogers1978finite}{DBLP:journals/neco/KearnsR99}.

\begin{definition}
    \label{definition:hypothesis-stability}
    Let $m,k \in \bbN$, $k < m$, $\alpha,\beta \in [0,1]$. Let $\cX$ be a set, let $A: ~ \left(\cX \times \pmo\right)^* \to \pmo^\cX$ be a learning rule, and let $\bbD \subseteq \distribution{\cX \times \pmo}$. We say that \ul{$A$ is $(\alpha,\beta, m, k)$-hypothesis stable with respect to $\bbD$} if 
    \[
        \forall \cD \in \bbD: ~ \PPPunder{
            \substack{
                S_1 \sim \cD^{m-k}
                \\
                S_2 \sim \cD^{k}
            }
        }{
            \dist{\cD_\cX}{A(S_1), A(S_1 \circ S_2)} \leq \alpha
        } \geq 1-\beta,
    \]
    where $\cD_\cX$ is the marginal of $\cD$ on $\cX$, $\dist{\cP}{f, g} = \PPP{x \sim \cP}{f(x) \neq g(x)}$, and $\circ$ denotes concatenation.
\end{definition}

\begin{definition}
    \label{definition:loss-stability}
    In the notation of \cref{definition:hypothesis-stability}, we say that \ul{$A$ is $(\alpha,\beta, m, k)$-loss stable with respect to $\bbD$} if 
    $
        \forall \cD \in \bbD: ~ \PPP{
                S_1 \sim \cD^{m-k},
                S_2 \sim \cD^{k}
        }{
            \Big|
                \Loss{\cD}{A(S_1)} - \Loss{\cD}{A(S_1 \circ S_2)} 
            \Big|
                \leq \alpha
        } \geq 1-\beta
    $.
\end{definition}

\begin{theorem}
    \label{theorem:stablity-implies-estimability}
    Let $k \in \bbN$ and $\alpha_0,\beta_0\in (0,1)$ such that $k \geq \OmegaOf{\log(1/\beta_0)/\alpha_0^2}$. Let $A$ be a learning rule that is $(\alpha_1,\beta_1, m, k)$-hypothesis stable or loss stable with respect to $\bbD$ (as in \cref{definition:hypothesis-stability,definition:loss-stability}). Then $(A,\bbD)$ is $(\varepsilon = \alpha_0+\alpha_1,\delta=\beta_0+\beta_1,m)$-uniformly estimable.
\end{theorem}

\begin{proof}
    If $(A,\bbD)$ is $(\alpha,\beta, m, k)$-hypothesis stable, then in particular $(A,\bbD)$ is also $(\alpha,\beta, m, k)$-loss stable. Hence, it suffices to prove the claim for the case of loss stability. We construct a uniform estimator $\estimator$ as follows. Given a sample $S \in \cZ^m$ for $\cZ = \left(\cX \times \pmo \right)$, let $S_1\circ S_2 = S$ be the partition of $S$ such that $S_1 \in \cZ^{m-k}$ and $S_2 \in \cZ^k$. Take $\estimator(S) = \Loss{S_2}{A(S_1)}$.

    By the triangle inequality,
    \begin{align*}
        |\estimator(S) - \Loss{\cD}{A(S)}| 
        &\leq
        \left|\estimator(S) - \Loss{\cD}{A(S_1)}\right| + \left|\Loss{\cD}{A(S_1)} - \Loss{\cD}{A(S)}\right|,
    \end{align*}
    so
    % By Hoeffding's inequality, for any distribution $\cD$, $\PPP{S \sim \cD^m}{}$
    \begin{align*}
        \PPP{S \sim \cD^m}{
            |\estimator(S) - \Loss{\cD}{A(S)}| > \varepsilon
        }
        &\leq 
        \PP{
            \begin{array}{l}
            |\Loss{S_2}{A(S_1)} - \Loss{\cD}{A(S_1)}| > \alpha_0
            ~ \lor
            \\
            |\Loss{\cD}{A(S_1)} - \Loss{\cD}{A(S)}| > \alpha_1
            \end{array}
        }
        \\
        &\leq 
        \PP{
            |\Loss{S_2}{A(S_1)} - \Loss{\cD}{A(S_1)}| > \alpha_0
        }
        \\ 
        & \qquad + \PP{
            |\Loss{\cD}{A(S_1)} - \Loss{\cD}{A(S)}| > \alpha_1
        }
        \\
        &\leq 
        \beta_0 + \beta_1 = \delta,
    \end{align*}
    where the final step follows from Hoeffding's inequality, the choice of $k$, and the stability of~$A$.
\end{proof}

Hence, stability is a sufficient condition for estimability. We remark that it is not a necessary condition. For instance, a learning rule selected at random as in \cref{example:random-algorithm} most likely is estimable (because it has high loss for any distribution), but not hypothesis stable (since for each possible input sample, it outputs a different hypothesis that was chosen at random). To see that loss stability is also not necessary for estimability, fix a degenerate distribution $\cD$ such that $\cD((x^*, 1))=1$ for some $x^*$, and consider an algorithm $A$ that for samples of size $m$ outputs the constant hypothesis $h_1(x) = 1$, and for samples of size $m-k$ outputs the constant hypothesis $h_0(x) = 0$. $A$ is perfectly estimable with respect to $\{\cD\}$, but it is not loss stable. 

%two estimable algorithms $A_m$ and $A_{m-k}$ that take training sets of size $m$ and $m-k$ samples respectively, and respectively achieve very low and very high loss for $\cD$. Take $A$ to be an algorithm that simulates $A_m$ on samples of size $m$ and $A_{m-k}$ on samples of size $m-k$. Then $A$ is estimable but is not loss stable.

One might object that \cref{theorem:stablity-implies-estimability} is of limited utility, because it is hard to check whether a given  algorithm is hypothesis stable or loss stable. Our response to this criticism is that in practice, it is quite easy to check whether an algorithm is loss (or hypothesis) stable with respect to a particular population distribution -- and indeed we do so in our experiments (see \cref{section:experiments}).

The process for estimating loss stability is simple: take a set $S$ of $m$ i.i.d.\ labeled samples from the population distribution. Randomly choose a subset $S'$ of size $m-k$. Execute the learning algorithm twice, once with training set $S$ to produce a hypothesis $h$, and another time with training set $S'$ to produce a hypothesis $h'$. Use an additional validation set to estimate the difference in population loss between $h$ and $h'$. Repeating this process a number of times and taking an average gives a good estimate of the $(m, k)$-loss stability. A similar process can be used to estimate hypothesis stability. Simply measure the disagreement between $h$ and $h'$ on the validation set (note that in this case, the validation set can be unlabeled, which is an advantage when labeling data is expensive).

\section{A Simple Characterization}

The following definition is a variant of Definition~\ref{definition:average-estimability}. Such a variant allows us to have a simple characterization of estimability in \cref{fact:tautology}. Namely, to understand whether an algorithm is estimable with respect to a set of distributions, one can examine the quantity $\EEE{\cD \sim \uniform{\bbD},S \sim \cD^m}{\operatorname{var}(L_{\mathcal{D}}({A}(S)) \mid S)}$. %(over the probability function ${D \sim \mathcal{U}(\mathcal{D}), S \sim \mathcal{D}^m}$) that appears in \cref{fact:tautology}. 

\label{section:characterization}

\begin{definition}\label{def:tau}
     Let $\mathbb{D}$ be a set of distributions and let $A$ be a learning algorithm. We say that $A$ is $(\varepsilon,m)$-estimable in $\ell_2$ with respect to $\mathbb{D}$, if there exists an estimator $\estimator$ such that
\[
     \EEE{\cD \sim \uniform{\bbD},S \sim \cD^m}{ (\estimator(S) - L_{\mathcal{D}}({A}(S)))^2} \leq \varepsilon
\]
\end{definition}

We remark that for bounded loss functions, one can move between Definition~\ref{def:tau} and Definition~\ref{definition:average-estimability} using Markov's inequality. Furthermore, although the characterization in the following theorem is simple, it might provide a technical condition that will be useful for future work.

\begin{fact}\label{fact:tautology}
     $A$ is $(\varepsilon,m)$-estimable in $\ell_2$ with respect to $\bbD$ if and only if
    \[
     \EEE{\cD \sim \uniform{\bbD},S \sim \cD^m}{\operatorname{var}(L_{\mathcal{D}}({A}(S)) \mid S)} \leq \varepsilon.
    \]
\end{fact}

The proof of \cref{fact:tautology} appears in \cref{section:proof-of-tautology}.

\addcontentsline{toc}{section}{References}

\printbibliography

@article{hoeffding1963probability,
  author       = {Hoeffding, Wassily},
  publisher    = {Taylor \& Francis},
  url          = {https://doi.org/10.2307/2282952},
  date         = {1963},
  doi          = {doi.org/10.2307/2282952},
  journaltitle = {Journal of the American Statistical Association},
  pages        = {13--30},
  title        = {Probability Inequalities for Sums of Bounded Random Variables},
}

@article{bousquet2002stability,
  author       = {Bousquet, Olivier and Elisseeff, André},
  publisher    = {JMLR. org},
  date         = {2002},
  journaltitle = {The Journal of Machine Learning Research},
  pages        = {499--526},
  title        = {Stability and generalization},
  volume       = {2},
}

@article{info_stab2,
  author       = {Rammal, Mohamad Rida and Achille, Alessandro and Golatkar, Aditya and Diggavi, Suhas and Soatto, Stefano},
  date         = {2022},
  journaltitle = {Advances in Neural Information Processing Systems},
  pages        = {10179--10190},
  title        = {On leave-one-out conditional mutual information for generalization},
  volume       = {35},
}

@inproceedings{info_stab1,
  author       = {Haghifam, Mahdi and Moran, Shay and Roy, Daniel M and Dziugiate, Gintare Karolina},
  organization = {IEEE},
  booktitle    = {2022 IEEE International Symposium on Information Theory (ISIT)},
  date         = {2022},
  pages        = {2487--2492},
  title        = {Understanding generalization via leave-one-out conditional mutual information},
}

@article{stab_Lei2022StabilityAG,
  author       = {Lei, Yunwen and Jin, Rong and Ying, Yiming},
  url          = {https://api.semanticscholar.org/CorpusID:252383365},
  date         = {2022},
  journaltitle = {ArXiv},
  title        = {Stability and Generalization Analysis of Gradient Methods for Shallow Neural Networks},
  volume       = {abs/2209.09298},
}

@article{stable_elisseeff05a,
  author       = {Elisseeff, Andre and Evgeniou, Theodoros and Pontil, Massimiliano},
  url          = {http://jmlr.org/papers/v6/elisseeff05a.html},
  date         = {2005},
  journaltitle = {Journal of Machine Learning Research},
  number       = {3},
  pages        = {55--79},
  title        = {Stability of Randomized Learning Algorithms},
  volume       = {6},
}

@inproceedings{johnson1984extensions,
  author       = {Johnson, William B. and Lindenstrauss, Joram},
  organization = {American Mathematical Society},
  booktitle    = {Conference on Modern Analysis and Probability},
  date         = {1984},
  doi          = {10.1090/conm/026/737400},
  pages        = {189--206},
  title        = {Extensions of {L}ipschitz mappings into a {H}ilbert space},
  volume       = {26},
}

@article{DBLP:journals/jcss/Achlioptas03,
  author       = {Achlioptas, Dimitris},
  url          = {https://doi.org/10.1016/S0022-0000(03)00025-4},
  bibsource    = {dblp computer science bibliography, https://dblp.org},
  date         = {2003},
  doi          = {10.1016/S0022-0000(03)00025-4},
  journaltitle = {Journal of Computer and System Sciences},
  number       = {4},
  pages        = {671--687},
  title        = {Database-friendly random projections: {J}ohnson--{L}indenstrauss with binary coins},
  volume       = {66},
}

@book{DBLP:books/cu/BV2014,
  author    = {Boyd, Stephen P. and Vandenberghe, Lieven},
  publisher = {Cambridge University Press},
  url       = {https://web.stanford.edu/\%7Eboyd/cvxbook/},
  bibsource = {dblp computer science bibliography, https://dblp.org},
  date      = {2014},
  doi       = {10.1017/CBO9780511804441},
  isbn      = {978-0-521-83378-3},
  title     = {Convex Optimization},
}

@inproceedings{DBLP:conf/iclr/JiangNMKB20,
  author    = {Jiang, Yiding and Neyshabur, Behnam and Mobahi, Hossein and Krishnan, Dilip and Bengio, Samy},
  publisher = {OpenReview.net},
  url       = {https://openreview.net/forum?id=SJgIPJBFvH},
  bibsource = {dblp computer science bibliography, https://dblp.org},
  booktitle = {The 8th International Conference on Learning Representations, {ICLR} 2020, Addis Ababa, Ethiopia, April 26-30, 2020},
  date      = {2020},
  title     = {Fantastic Generalization Measures and Where to Find Them},
}

@inproceedings{gastpar2023fantastic,
  author    = {Gastpar, Michael and Nachum, Ido and Shafer, Jonathan and Weinberger, Thomas},
  url       = {https://openreview.net/pdf?id=NkmJotfL42},
  booktitle = {The 12th International Conference on Learning Representations, {ICLR} 2024},
  date      = {2024},
  title     = {Fantastic generalization measures are nowhere to be found},
}

@inproceedings{DBLP:conf/nips/DziugaiteDNRCWM20,
  author    = {Dziugaite, Gintare Karolina and Drouin, Alexandre and Neal, Brady and Rajkumar, Nitarshan and Caballero, Ethan and Wang, Linbo and Mitliagkas, Ioannis and Roy, Daniel M.},
  editor    = {Larochelle, Hugo and Ranzato, Marc'Aurelio and Hadsell, Raia and Balcan, Maria{-}Florina and Lin, Hsuan{-}Tien},
  url       = {https://proceedings.neurips.cc/paper/2020/hash/86d7c8a08b4aaa1bc7c599473f5dddda-Abstract.html},
  bibsource = {dblp computer science bibliography, https://dblp.org},
  booktitle = {Advances in Neural Information Processing Systems 33: Annual Conference on Neural Information Processing Systems 2020, NeurIPS 2020, December 6-12, 2020, virtual},
  date      = {2020},
  title     = {In search of robust measures of generalization},
}

@inproceedings{DBLP:conf/aistats/ViallardEHMZ24,
  author    = {Viallard, Paul and Emonet, Rémi and Habrard, Amaury and Morvant, Emilie and Zantedeschi, Valentina},
  editor    = {Dasgupta, Sanjoy and Mandt, Stephan and Li, Yingzhen},
  publisher = {PMLR},
  url       = {https://proceedings.mlr.press/v238/viallard24a.html},
  bibsource = {dblp computer science bibliography, https://dblp.org},
  booktitle = {International Conference on Artificial Intelligence and Statistics, 2-4 May 2024, Palau de Congressos, Valencia, Spain},
  date      = {2024},
  pages     = {3007--3015},
  series    = {Proceedings of Machine Learning Research},
  title     = {Leveraging PAC-Bayes Theory and Gibbs Distributions for Generalization Bounds with Complexity Measures},
  volume    = {238},
}

@article{blake2006properties,
  author       = {Blake, Ian F and Studholme, Chris},
  url          = {http://www.cs.toronto.edu/~cvs/coding/random_report.pdf},
  date         = {2006},
  journaltitle = {Unpublished report},
  title        = {Properties of random matrices and applications},
}

@article{DBLP:journals/neco/KearnsR99,
  author       = {Kearns, Michael J. and Ron, Dana},
  url          = {https://doi.org/10.1162/089976699300016304},
  bibsource    = {dblp computer science bibliography, https://dblp.org},
  date         = {1999},
  doi          = {10.1162/089976699300016304},
  journaltitle = {Neural Comput.},
  number       = {6},
  pages        = {1427--1453},
  title        = {Algorithmic Stability and Sanity-Check Bounds for Leave-One-Out Cross-Validation},
  volume       = {11},
}

@article{rogers1978finite,
  author       = {Rogers, William H and Wagner, Terry J},
  publisher    = {JSTOR},
  date         = {1978},
  journaltitle = {The Annals of Statistics},
  pages        = {506--514},
  title        = {A finite sample distribution-free performance bound for local discrimination rules},
}

@inproceedings{Haghifam2022,
  author    = {Haghifam, Mahdi and Moran, Shay and Roy, Daniel M. and Dziugiate, Gintare Karolina},
  booktitle = {2022 IEEE International Symposium on Information Theory (ISIT)},
  date      = {2022},
  doi       = {10.1109/ISIT50566.2022.9834400},
  pages     = {2487--2492},
  title     = {Understanding Generalization via Leave-One-Out Conditional Mutual Information},
}

@inproceedings{dziugaite2021role,
  author       = {Dziugaite, Gintare Karolina and Hsu, Kyle and Gharbieh, Waseem and Arpino, Gabriel and Roy, Daniel},
  organization = {PMLR},
  booktitle    = {International Conference on Artificial Intelligence and Statistics},
  date         = {2021},
  pages        = {604--612},
  title        = {On the role of data in PAC-Bayes bounds},
}

@inproceedings{negrea2020defense,
  author       = {Negrea, Jeffrey and Dziugaite, Gintare Karolina and Roy, Daniel},
  organization = {PMLR},
  booktitle    = {International Conference on Machine Learning},
  date         = {2020},
  pages        = {7263--7272},
  title        = {In defense of uniform convergence: Generalization via derandomization with an application to interpolating predictors},
}

@article{nagarajan2019uniform,
  author       = {Nagarajan, Vaishnavh and Kolter, J Zico},
  date         = {2019},
  journaltitle = {Advances in Neural Information Processing Systems},
  title        = {Uniform convergence may be unable to explain generalization in deep learning},
  volume       = {32},
}

@article{bartlett2021failures,
  author       = {Bartlett, Peter L and Long, Philip M},
  publisher    = {JMLRORG},
  date         = {2021},
  journaltitle = {The Journal of Machine Learning Research},
  number       = {1},
  pages        = {9297--9311},
  title        = {Failures of model-dependent generalization bounds for least-norm interpolation},
  volume       = {22},
}

@article{xu2017information,
  author       = {Xu, Aolin and Raginsky, Maxim},
  date         = {2017},
  journaltitle = {Advances in Neural Information Processing Systems},
  title        = {Information-theoretic analysis of generalization capability of learning algorithms},
  volume       = {30},
}

@inproceedings{bassily2018learners,
  author       = {Bassily, Raef and Moran, Shay and Nachum, Ido and Shafer, Jonathan and Yehudayoff, Amir},
  organization = {PMLR},
  booktitle    = {Algorithmic Learning Theory},
  date         = {2018},
  pages        = {25--55},
  title        = {Learners that use little information},
}

@article{zhang2023lower,
  author       = {Zhang, Peiyuan and Teng, Jiaye and Zhang, Jingzhao},
  date         = {2023},
  journaltitle = {arXiv preprint arXiv:2303.10758},
  title        = {Lower Generalization Bounds for GD and SGD in Smooth Stochastic Convex Optimization},
}

@inproceedings{colin,
  author    = {Abbe, Emmanuel and Sandon, Colin},
  editor    = {Larochelle, H. and Ranzato, M. and Hadsell, R. and Balcan, M.F. and Lin, H.},
  publisher = {Curran Associates, Inc.},
  url       = {https://proceedings.neurips.cc/paper_files/paper/2020/file/e7e8f8e5982b3298c8addedf6811d500-Paper.pdf},
  booktitle = {Advances in Neural Information Processing Systems},
  date      = {2020},
  pages     = {20061--20072},
  title     = {On the universality of deep learning},
  volume    = {33},
}

@inproceedings{nikolakakis2023beyond,
  author    = {Nikolakakis, Konstantinos and Haddadpour, Farzin and Karbasi, Amin and Kalogerias, Dionysios},
  url       = {https://openreview.net/forum?id=pOyi9KqE56b},
  booktitle = {The Eleventh International Conference on Learning Representations},
  date      = {2023},
  title     = {Beyond Lipschitz: Sharp Generalization and Excess Risk Bounds for Full-Batch {GD}},
}

@article{harutyunyan2021information,
  author       = {Harutyunyan, Hrayr and Raginsky, Maxim and Ver Steeg, Greg and Galstyan, Aram},
  date         = {2021},
  journaltitle = {Advances in Neural Information Processing Systems},
  pages        = {24670--24682},
  title        = {Information-theoretic generalization bounds for black-box learning algorithms},
  volume       = {34},
}

@article{hellstrom2022new,
  author       = {Hellström, Fredrik and Durisi, Giuseppe},
  date         = {2022},
  journaltitle = {Advances in Neural Information Processing Systems},
  pages        = {10108--10121},
  title        = {A new family of generalization bounds using samplewise evaluated CMI},
  volume       = {35},
}

@article{wang2023tighter,
  author       = {Wang, Ziqiao and Mao, Yongyi},
  date         = {2023},
  journaltitle = {arXiv preprint arXiv:2302.02432},
  title        = {Tighter Information-Theoretic Generalization Bounds from Supersamples},
}

@article{EspositoGI:2021,
  author       = {Esposito, Amedeo Roberto and Gastpar, Michael and Issa, Ibrahim},
  date         = {2021},
  doi          = {10.1109/TIT.2021.3085190},
  journaltitle = {IEEE Transactions on Information Theory},
  number       = {8},
  pages        = {4986--5004},
  title        = {Generalization Error Bounds via {R}ényi-, f-Divergences and Maximal Leakage},
  volume       = {67},
}

@inproceedings{pmlr-v195-issa23a,
  author    = {Issa, Ibrahim and Esposito, Amedeo Roberto and Gastpar, Michael},
  editor    = {Neu, Gergely and Rosasco, Lorenzo},
  publisher = {PMLR},
  url       = {https://proceedings.mlr.press/v195/issa23a.html},
  booktitle = {Proceedings of Thirty Sixth Conference on Learning Theory},
  date      = {2023},
  pages     = {4952--4976},
  series    = {Proceedings of Machine Learning Research},
  title     = {Generalization Error Bounds for Noisy, Iterative Algorithms via Maximal Leakage},
  volume    = {195},
}

@inproceedings{IssaEG:2019,
  author    = {Issa, Ibrahim and Esposito, Amedeo Roberto and Gastpar, Michael},
  booktitle = {2019 IEEE International Symposium on Information Theory (ISIT)},
  date      = {2019},
  doi       = {10.1109/ISIT.2019.8849834},
  pages     = {582--586},
  title     = {Strengthened Information-theoretic Bounds on the Generalization Error},
}

@book{grimmett2020probability,
  title={Probability and random processes},
  author={Grimmett, Geoffrey and Stirzaker, David},
  year={2020},
  publisher={Oxford university press}
}

\addcontentsline{toc}{section}{Appendices}

\appendix

\section{Proof of Theorem~\ref{theorem:vc-class-estimability-bound}}

\label{appendix:proof-vc}

Recall the definition of nearly-orthogonal functions (\cref{definition:eta-orthogonal}). 
The proof of \cref{theorem:vc-class-estimability-bound} uses a corollary of the  Johnson--Lindenstrauss lemma (\cref{theorem:johnson-lindenstrauss}), which states that random vectors in a high dimensional space are nearly orthogonal, as follows.\footnote{It is also possible to prove a similar claim by directly using concentration of measure (e.g., Hoeffding's inequality), without using the Johnson--Lindenstrauss lemma.}

\begin{claim}
	\label{claim:random-functions-are-orthogonal}
	Let $\varepsilon \in (0,1/2)$, and let $d,n \in \bbN$ such that
	% \[
	% 	d \geq \frac{18}{\varepsilon^2/2 - \varepsilon^3/3}\ln n.	
	% \]
	\[
		n \leq \expf{d\varepsilon^2 / 54}.	
	\]
	Let $\cU = \uniform{\pmo^{[d]}}$ be the uniform distribution over functions $[d] \to \pmo$, and consider a random sequence $F$ of functions $F_1,\dots,F_n$ sampled independently from $\cU$. Then 
	\[
		\PPP{F \sim \cU^n}{F \in \orthfull{\varepsilon}{[d]}} \geq 0.99.
	\]
\end{claim}

\begin{proof}[Proof of \cref{claim:random-functions-are-orthogonal}]
	If $n = 1$ there is nothing to prove, so we assume $n \geq 2$. 
	Let $R \sim \uniform{\pmo^{d \times n}}$ be a $d \times n$ matrix with entries in $\pmo$ chosen independently and uniformly at random. In particular, for each $i \in [n]$, the $i$-th column of $R$ is a vector of $d$ numbers in $\pmo$ chosen independently and uniformly at random. Hence, using $e_1,\dots,e_n$ to denote the standard basis of $\bbR^n$, we identify the vector $Re_i$, which is the $i$-th column of $R$, with the random function $F_i: ~ [d] \to \pmo$.

	Recall that for vectors $u,v \in \bbR^d$,
	\begin{align*}
		\|u - v\|_2^2 = \iprod{u-v,u-v} = \|u\|_2^2 -2\iprod{u,v} + \|v\|_2^2,
	\end{align*}
	so
	\begin{align}
		\label{eq:inner-prod-to-norm}
		\iprod{u,v} = \frac{\|u\|_2^2 + \|v\|_2^2 - \|u - v\|_2^2}{2}.
	\end{align}

	Invoking \cref{theorem:johnson-lindenstrauss} with $s=n$, $\beta=7$, $V = \{e_1,\dots,e_n\} \subseteq \bbR^n$, and $d,n,\varepsilon$ as in the claim statement implies that
	\begin{align}
		\label{eq:JL-for-standard-basis}
		\PPPunder{R \sim \uniform{\pmo^{d \times n}}}{
			\begin{array}{l}
				\forall i,j \in [n], i \neq j: 
				\\[0.5em]
				\quad (1-\varepsilon)\cdot2 
				\leq 
				\left\|
					\frac{1}{\sqrt{d}}Re_i-\frac{1}{\sqrt{d}}Re_j
				\right\|_2^2 
				\leq 
				(1+\varepsilon)\cdot2
			\end{array}
		} \geq 1-\frac{1}{n^{\beta}}.		
	\end{align}
    Hence, with probability at least $1-1/n^{\beta} \geq 1-1/2^7 \geq 0.99$ over the choice of $F$, every distinct $i,j \in [n]$ satisfy
	\vspace*{-0.7em}
	\begin{align*}
		\left|\EEE{x \sim \uniform{[d]}}{F_i(x)F_j(x)}\right|
		&=
		\left|\frac{1}{d}\sum_{x \in [d]}F_i(x)F_j(x)\right|
		=
		\left|\frac{1}{d}\iprod{Re_i,Re_j}\right|
		\tagexplain{Identifying $F_i$ with $Re_i$}
		\\
		&= 
		\left|\frac{\|Re_i\|_2^2 + \|Re_j\|_2^2 - \|Re_i - Re_i\|_2^2}{2d}\right|
		\tagexplain{By \cref{eq:inner-prod-to-norm}}
		\\
		&= 
		\left|1 - \frac{1}{2}\left\|\frac{1}{\sqrt{d}}Re_i - \frac{1}{\sqrt{d}}Re_i\right\|_2^2\right|
		\\
		&\leq
		\varepsilon,
		\tagexplain{By \cref{eq:JL-for-standard-basis}}
	\end{align*}
	as desired.
\end{proof}

\begin{proof}[Proof of \cref{theorem:vc-class-estimability-bound}]
	Fix an $\cH$-shattered set $\cX_d \subseteq \cX$ with cardinality $|\cX_d|=d$, and for each $f: ~ \cX_d \to \pmo$ let $\cD_f = \uniform{\{(x,f(x)): ~ x \in \cX_d\}}$. 
	Note that the distributions $\cD_f$ are $\cH$-realizable.
	We will show that there exists a collection $\bbD = \{\cD_f: f \in \cF\}$ that satisfies \cref{eq:vc-class-not-estimable}, where $\cF \subseteq \pmo^{\cX_d}$ is a set of $k=2^m+1$ functions.

	Consider the following experiment:
	\begin{enumerate}
		\item{
			\label{item:sample-cF}
			Sample a sequence of functions $G = (G_1,\dots,G_k)$ independently and uniformly at random from $\pmo^{\cX_d}$.
		}
		\item{
			\label{item:sample-F} 	
			Sample a function $F$ uniformly from $G$.
		}
		\item{
			\label{item:sample-X}
			Sample a sequence of points $X = (X_1,\dots,X_m)$ independently and uniformly at random from $\cX_d$. ($X$ is sampled independently of $(G,F)$.)
			%\footnote{\cref{item:sample-X} is performed independently of \cref{item:sample-cF}, and \cref{item:sample-F} is performed independently of \cref{item:sample-X}.}
		}
		\item{
			For each $i \in [m]$, let $Y_i = F(X_i)$, let $Y = (Y_1,\dots, Y_m)$, and let $S = \Bigl((X_1,Y_1),\dots, (X_m,Y_m)\Bigr)$.
		}
	\end{enumerate}
	Let $\cP$ be the joint distribution of $(G, F, X, Y, S)$. Consider the following events:
	\begin{itemize}
		\item{
			$\cE_1 = \left\{
				G \in \orthfull{\varepsilon}{\cX_d}
			\right\}$ for $\varepsilon = 2/d^{1/4}$. 
			By \cref{claim:random-functions-are-orthogonal} and the choice of $k$, $\cPP{\cE_1} \geq 0.99$ for $d$ large enough.\footnote{We choose $d_0 \in \bbN$ to be the universal constant such that this inequality holds for all integers $d \geq d_0$ and all $m \leq \sqrt{d}/10$.}
		}
		\item{
			$\cE_2 = \big\{
				\left|
					\{X_1,\dots,X_m\}
				\right| = m
			\big\}$. 
			By \cref{claim:birthday-paradox} and the choice of $m$, $\cPP{\cE_2} \geq 0.99$.
		}
		\item{
			$\cE_3 = \bigl\{\left|G_S\right| = 2\bigr\}$. $\cPP{\cE_3 \: \vert \: \cE_2} \geq 1/e$. To see this, note that each function $G_i \in G \setminus \{F\}$ is chosen independently of $F$. Hence, the probability that a function $G_i$ agrees with $F$ on the $m$ distinct samples in $X$ (i.e., the probability that $G_i(X_j) = F(X_j)$ for all $j \in [m]$, given $\cE_2$) is $p = 2^{-m}$. The functions in $G$ are chosen independently, so the number $T$ of functions in $G \setminus \{F\}$ that agree with $F$ on $m$ distinct samples has a binomial distribution $T \sim \Binomial{k-1, p}$. So
			\begin{align*}
				\PP{T = 1} 
				&=
				(k-1)\cdot p \cdot (1-p)^{k-2}
				=
				(1-p)^{k-2}
				\\
				&\geq
				\left(e^{-\frac{p}{1-p}}\right)^{k-2}
				\tagexplain{$\forall p < 1: ~ 1-p \geq e^{-p/(1-p)}$}
				\\
				&
				=
				1/e.
			\end{align*}
		}
	\end{itemize}
	Let $\cE = \cE_1 \cap \cE_3$. Combining the above bounds yields
	\begin{align*}
		\cPP{\cE}
		&= 
		\cPP{\cE_1 \cap \cE_3} 
		\\
		&\geq
		\cPP{\cE_3}-\cPP{\cE_1^C}
		\\
		&\geq
		\cPP{\cE_3 \: \vert \: \cE_2}\cdot\cPP{\cE_2} -\cPP{\cE_1^C}
		\\
		&\geq
		0.99 \cdot 1/e - 0.01 
		> 1/3.
	\end{align*}
	By an averaging argument, this implies that there exists $\cF \subseteq \pmo^{\cX_d}$ such that $\cF \in \orthfull{\varepsilon}{\cX_d}$ for $\varepsilon = 2/d^{1/4}$ and 
	\begin{equation}
		\label{eq:GS-eq-2}
		\cPP{\left|G_S\right| = 2 \: \vert \: G = \cF} \geq 1/3.
	\end{equation}
	Fix this $\cF$, and let $A$ be an $\cF$-interpolating learning rule. From the technical lemma (\cref{lemma:technical}), there exists a collection of $\cF$-realizable distributions $\bbD \subseteq \distribution{\cX_d \times \pmo}$ such that for any estimator $\estimator: \: (\cX \times \pmo)^m \to [0,1]$ that may depend on $\bbD$ and $A$, 
	\begin{align*}
		\PPP{
			\substack{\cD \sim \uniform{\bbD} 
			\\ 
			S \sim \cD^m}
		}{
			\bigl|\estimator(S) - \Loss{\cD}{A(S)}\bigr| \geq \frac{1}{4}-\frac{\varepsilon}{4}
		} 
		&\geq 
		\frac{1}{2}\cdot \PPP{
			\substack{\cD \sim \uniform{\bbD} 
			\\ 
			S \sim \cD^m}
		}{|\cF_S| = 2}
		\\
		&\geq 
		\frac{1}{2} \cdot \frac{1}{3}
		=
		\frac{1}{6}
		,
		\tagexplain{By \cref{eq:GS-eq-2}}
	\end{align*}

	as desired.
\end{proof}

\section{Proof of Theorem~\ref{theorem:orthogonal-functions}}

\label{appendix:proof-orthogonal}

\begin{proof}[Proof of \cref{theorem:orthogonal-functions}]
	We take $\bbD = \left\{\cD_f: \: f \in \cF\right\}$ where $\cD_f = \uniform{\{(x,f(x)): ~ x \in \cX\}}$. 
	Fix a function $f^* \in \cF$, let $S \sim (\cD_{f^*}\!)^m$, and consider the random variable $Z = |\cF_S|$.
	We bound the expectation and variance of $Z$, and then show a lower bound on the probability that $Z \in \{2,3\}$.

	Let $S = \big((X_1,Y_1),\dots,(X_m,Y_m)\big)$ and $X = \{X_1,\dots,X_m\}$, and let $E$ denote the event in which $|X| = m$ (i.e., $S$ is collision-free).
	For each $f \in \cF$, let $Z_f = \indicator{\forall i \in [m]: \: f(X_i) = Y_i}$, so that $Z = \sum_{f \in \cF}Z_f$.

	\begingroup
	\allowdisplaybreaks
	\begin{align}
		\EEEunder{
			S \sim (\cD_{f^*}\!)^m
		}{
			Z \: | \: E
		}
		&=
		\EE{\sum_{f \in \cF}Z_f \: \Big| \: E}
		\nonumber
		\\
		&=
		1 + \sum_{\substack{f \in \cF
		\\
		f \neq f^*}}
		\PP{\forall i \in [m]: \: f(X_i) = Y_i
		\: \big| \: E}
		\tagexplain{$Z_F = 1$}
		\nonumber
		\\
		&\leq
		1 + 2^m\cdot
			\left(
				\frac{1}{2} + \frac{1}{2}\cdot\frac{1}{1000m}
			\right)^m
		\tagexplainhspace{By \cref{fact:orthogonal-loss}}{-2em}
		\nonumber
		\\
		&\leq
		1 + e^{1/1000} < 2.002.
		\label{eq:EZ-upper}
		\\
		~
		\nonumber\\
		\EEEunder{
			S \sim (\cD_{f^*}\!)^m
		}{
			Z \: | \: E
		} &\geq 1 + 2^m\cdot
		\left(
			\frac{1}{2} - \frac{1}{2}\cdot\frac{1}{1000m}
		\right)^m
		\tagexplainhspace{By \cref{fact:orthogonal-loss}}{-2em}
		\nonumber
		\\
		&\geq
		1 + e^{-1/500}.
		\tagexplainhspace{$1-x \geq e^{-x/(1-x)}$}{-4em}
		\label{eq:EZ-lower}
		\\
		~
		\nonumber\\
		\EEEunder{
			S \sim (\cD_{f^*}\!)^m
		}{
			Z^2 \: | \: E
		}
		&=
		\EE{
			\left(\sum_{f \in \cF}Z_f\right)\left(\sum_{g \in \cF} Z_g\right) \: \Big| \: E
		}
		\nonumber
		\\
		&=
		\EE{
			\left(1 + \sum_{\substack{f \in \cF \\ f \neq f^*}}Z_f\right)\left(1 + \sum_{\substack{g \in \cF \\ g \neq f^*}} Z_g\right) \: \Big| \: E
		}
		\tagexplain{$Z_{f^*} = 1$}
		\nonumber
		\\
		&=
		\EE{
			1+
			2\sum_{\substack{f \in \cF
			\\
			f \neq f^*}}
			Z_f
			+
			\sum_{\substack{f \in \cF
			\\
			f \neq f^*}}
			\sum_{
				\substack{g \in \cF
				\\
				g \neq f^*}
			}
			Z_f Z_g
			\: \Big| \: E
		}
		\nonumber
		\\
		&=
		\EE{
			1+
			3\sum_{\substack{f \in \cF
			\\
			f \neq f^*}}
			Z_f
			+
			\sum_{\substack{f,g \in \cF \setminus \{f^*\}
			\\
			f \neq g}}
			Z_f Z_g
			\: \Big| \: E
		}
		\nonumber
		\\
		&=
			1
			+
			3\left(\EE{
				Z \: | \: E
			}-1\right)
			+
			\sum_{\substack{f,g \in \cF \setminus \{f^*\}
			\\
			f \neq g}}
			\EE{
				Z_f Z_g
				\: \Big| \: E
			}.
			\label{eq:Z-sqaured-upper}
		\\
		~
		\nonumber\\
		\sum_{\substack{f,g \in \cF \setminus \{f^*\}
			\\
			f \neq g}}
			\EE{
				Z_f Z_g
				\: \Big| \: E
			}
		&=
		\sum_{\substack{f,g \in \cF \setminus \{f^*\}
			\\
			f \neq g}}
			\PP{\forall i \in [m]: \: 
				f(X_i) = g(X_i) = f^*(X_i)
			\: \big| \: E}
			\nonumber
		\\
		&\leq
		2^{2m}
		\cdot
		\left(\frac{1}{4}+\frac{3}{4}\cdot\frac{1}{1000m}\right)^m
		\tagexplainhspace{By \cref{claim:orthogonal-three-way-agreement}}{-4em}
		\nonumber
		\\
		&=
		\left(1+\frac{3}{1000m}\right)^m
		\nonumber
		\\
		&\leq 
		e^{3/1000}.
		\label{eq:Z-cross-term-upper}
	\end{align}
	\endgroup

	Combining \cref{eq:EZ-upper,eq:EZ-lower,eq:Z-sqaured-upper,eq:Z-cross-term-upper} yields
	\begin{align*}
		\Var{
			Z \: | \: E
		}
		&=
		\EE{Z^2 \: | \: E} - \left(\EE{Z \: | \: E}\right)^2
		\\
		&\leq
		1
			+
			3e^{1/1000}
			+
			e^{3/1000}
			-
			\left(1 + e^{-1/500}\right)^2
		\\
		&<
		1.02.
		\end{align*}
		By \cref{lemma:lp-concentration}, 
		\begin{align*}
			\PP{Z \in \{2, 3\} \: | \: E} \geq 1 - \frac{\Var{
			Z \: | \: E
			}}{2} \geq 0.49.
		\end{align*}
		\cref{claim:birthday-paradox} and $|\cX'| \geq 100m^2$ imply that $\PP{E} \geq 0.99$. Hence,
		\begin{align*}
			\PP{Z \in \{2, 3\}} 
			&\geq 
			\PP{E}\cdot\PP{Z \in \{2, 3\} \: | \: E}
			\geq 
			0.99 \cdot 0.49 \geq 0.48.  
		\end{align*}
		Finally, invoking our technical lemma (\cref{lemma:technical}) yields
		\[
			\PPP{
				\substack{F \sim \uniform{\cF} \\ S \sim (\cD_F)^m}
			}{
				\bigl|\estimator(S) - \Loss{\cD_F}{A(S)}\bigr| \geq \frac{1}{4}-\frac{1}{4000m}
			} \geq \frac{\PP{Z \in \{2, 3\}}}{3} \geq 0.16,
		\]
		as desired.
\end{proof}

\section{Proof of Fact~\ref{fact:tautology}}

\label{section:proof-of-tautology}

\begin{proof}

The result that the minimum mean-square error (MMSE) estimator corresponds to the conditional expectation is a well-established theorem in probability theory (see, for instance, Section 7.9 in \textcite{grimmett2020probability}). For the sake of completeness, we present a proof of this result.

We will use the following simple claim.
\begin{claim}\label{cl:mean}
    Let $c_1,...,c_k,p_1,...,p_k \in \bbR$ such that $\sum_{i=1}^k p_i=1$, then 
    
    \[
        \operatorname{argmin}_{x\in\bbR}\sum_{i=1}^k p_i\cdot (x-c_i)^2= \sum_{i=1}^k p_i\cdot c_i.
    \] 
\end{claim}
The claim follows by taking the derivative of $\sum_{i=1}^k p_i\cdot (x-c_i)^2$ with respect to $x$ which yields the equation:

$\sum_{i=1}^{k} 2 p_i (x - c_i) = 0$ that implies $x = \sum_{i=1}^{k} p_i c_i$ since $\sum_{i=1}^{k} p_i=1$.

% Now, let $\estimator$ be any estimator for $A$.  

The following shows that the estimator  $\estimator^*(S) \coloneqq \bbE \left[  L_{\cD}({A}(S)) \mid S  \right]$ is optimal and the inequality follows from Claim~\ref{cl:mean}. Let $\estimator$ be any estimator for $A$.

\begin{flalign*}
    &
    \mathbb{E}_{\cD \sim \uniform{\bbD}, S \sim \cD^m} \left[ (\estimator(S) - L_{\cD}({A}(S)))^2 \right] 
    &
    \\
    &
    \qquad
    =  
    \sum_{S} \bbP(S)   \sum_{\cD\in \bbD}   \bbP(\cD |S)  \left( L_{\cD}({A}(S))-  \estimator(S) \right)^2  
    &
    \\
    &
    \qquad=  
    \bbE \left[  \sum_{\cD\in \bbD}  \bbP(\cD |S)  \left( L_{\cD}({A}(S))-  \estimator(S) \right)^2   \right] 
    &
    \\
    &
    \qquad\geq  
    \bbE \left[  \sum_{\cD\in \bbD}  \bbP(\cD |S)  \left( L_{\cD}({A}(S))-  \sum_{\cD\in \bbD}  \left[  \bbP(\cD |S)  L_{\cD}({A}(S))  \right] \right)^2   \right] 
    &
    \\
    &
    \qquad=   
    \bbE \left[  \sum_{\cD\in \bbD}  \bbP(\cD |S)  \left( L_{\cD}({A}(S))-  \estimator^*(S)  \right)^2   \right]
    &
    \\
    &
    \qquad=  
    \mathbb{E}_{\cD \sim \uniform{\bbD}, S \sim \cD^m} \left[ (\estimator^*(S) - L_{\cD}({A}(S)))^2 \right].
    &
\end{flalign*}

This means that $A$ is square loss $(\varepsilon,m)$-estimable with respect to $\bbD$ if and only if $\estimator^*$ can achieve $\varepsilon$ accuracy. It achieves such accuracy if and only if 
$\mathbb{E} \left[ \operatorname{var}(L_{\cD}({A}(S)) \mid S) \right] \leq \varepsilon
$. This follows by the following equalities that complete the proof.

% is square loss $(\varepsilon,m)$-estimable with respect to $\bbD$ we need

% The proof follows from the following.

\begin{align*}
    \mathbb{E} \left[ \operatorname{var}(L_{\cD}({A}(S)) \mid S) \right] &=  \bbE \left[ \bbE \left[  \left( L_{\cD}({A}(S))-  \bbE \left[ L_{\cD}({A}(S)) | S \right] \right)^2 \mid S  \right]  \right] \\
    &=  \bbE \left[  \sum_{\cD\in \bbD}     \bbP(\cD |S) \left( L_{\cD}({A}(S))-  \bbE \left[  L_{\cD}({A}(S)) \mid S  \right] \right)^2  \right]  \\
&=  \bbE \left[  \sum_{\cD\in \bbD}  \bbP(\cD |S)  \left( L_{\cD}({A}(S))-  \sum_{\cD\in \bbD}  \left[  \bbP(\cD |S)  L_{\cD}({A}(S))  \right] \right)^2   \right]  \\
&=  \bbE \left[  \sum_{\cD\in \bbD}  \bbP(\cD |S)  \left( L_{\cD}({A}(S))-   \estimator^*(S) \right)^2   \right] \\
&=      \mathbb{E}_{\cD \sim \uniform{\bbD}, S \sim \cD^m} \left[ (\estimator^*(S) - L_{\cD}({A}(S)))^2 \right]~\qedhere
\end{align*}
\end{proof}

\section{Technical Lemma for Inestimability}

\begin{lemma}
	\label{lemma:technical}
	Let $m \in \bbN$, let $\varepsilon > 0$, let $\cX$ be a finite set, let $\cF \subseteq \pmo^\cX$ such that $\cF \in \orthfull{\varepsilon}{\cX}$, and let $A: \: \left(\cX \times \pmo\right)^m \to \pmo^\cX$ be an $\cF$-interpolating learning rule. For each $f \in \cF$ let $\cD_f = \uniform{\left\{(x,f(x)): ~ x \in \cX\right\}}$, and for each $k \in \bbN$ let 
	\[
		p_k = \PPP{\substack{F \sim \uniform{\cF} \\ S \sim (\cD_F)^m}}{|\cF_S| = k}.	
	\]
	Then for any estimator $\estimator: \: (\cX \times \pmo)^m \to [0,1]$ that may depend on $A$, 
	\begin{equation*}
		\PPP{
			\substack{F \sim \uniform{\cF} \\ S \sim (\cD_F)^m}
		}{
			\bigl|\estimator(S) - \Loss{\cD_F}{A(S)}\bigr| \geq \frac{1}{4}-\frac{\varepsilon}{4}
		} \geq \sum_{k \in \left\{2,\dots,|\cF|\right\}} \frac{p_k}{k}.
	\end{equation*}
\end{lemma}

\begin{proof}
	Consider the following experiment:
	\begin{enumerate}
		\item{
			Sample a sequence of points $X = (X_1,\dots,X_m)$ independently and uniformly at random from $\cX$.
			% \footnote{\cref{item:sample-X} is performed independently of \cref{item:sample-cF}, and \cref{item:sample-F} is performed independently of \cref{item:sample-X}.}
		}
		\item{	
			Sample a function $F$ uniformly from $\cF$, independently of $X$.
		}
		\item{
			For each $i \in [m]$, let $Y_i = F(X_i)$, let $Y = (Y_1,\dots, Y_m)$, and let $S = \Bigl((X_1,Y_1),\dots, (X_m,Y_m)\Bigr)$.
		}
	\end{enumerate}
	Let $\cP$ be the joint distribution of $(X, F, Y, S)$. 
	Fix $k \in \left\{2,\dots,|\cF|\right\}$, and let 
	\[
		s = \big((x_1,y_1),\dots,(x_m,y_m)\big) \in \left(\cX \times \pmo\right)^m
	\]
	with $x = (x_1,\dots,x_m)$ and $y = (y_1,\dots,y_m)$ such that $|\cF_s| = k$. Denote $\cF_s = \{f_1,\dots,f_k\}$. 
	Then for any $i,j \in [k]$, $i \neq j$,
	\begin{align}
		\label{eq:S-given-F}
		\cPP{
			S = s ~ \vert ~ F = f_i
		}
		&=
		\cPP{
			X = x ~ \vert ~ F = f_i
		}
		\nonumber
		\\
		&=
		\cPP{
			X = x ~ \vert ~ F = f_j
		}
		\tagexplain{$X \bot F$}
		\nonumber
		\\
		&= \cPP{
			S = s ~ \vert ~ F = f_j
		}.
	\end{align}
	So,
	\begin{align}
		\label{eq:F-given-S}
		\cPP{
			F = f_i ~ \vert ~ S = s
		}
		&=
			\frac{\cPP{
				S = s ~ \vert ~ F = f_i
			}
			\cdot
			\cPP{
				F = f_i
			}
			}{
				\cPP{
					S = s
				}
			}
		\nonumber
		\\
		&=
			\frac{\cPP{
				S = s ~ \vert ~ F = f_j
			}
			\cdot
			\cPP{
				F = f_j
			}
			}{
				\cPP{
					S = s
				}
			}
			\tagexplain{By \cref{eq:S-given-F}, $F \sim \uniform{\cF}$}
		\nonumber
		\\
		&=
		\cPP{
			F = f_j ~ \vert ~ S = s
		},
	\end{align}
	Seeing as $\cPP{F \in \cF_s ~ | ~ S = s} = 1$, this implies that for all $i \in [k]$, $\cPP{
		F = f_i ~ \vert ~ S = s
	} = 1/k$.

	Because $A$ is $\cF$-interpolating, $A(s) \in \cF_s$. Without loss of generality, denote $A(s) = f_1$.
	From $\cF \in \orthfull{\varepsilon}{\cX}$ and \cref{fact:orthogonal-loss}, $\Loss{\cD_{f_i}}{f_j} \geq \frac{1}{2} - \frac{\varepsilon}{2}:= 2\alpha$ for all $i,j \in [k], i \neq j$. 
	Hence,
	\begin{align}
		\label{eq:p-zero-loss-eq-p-high-loss}
		\cPP{
			\Loss{\cD_F}{A(S)} = 0 ~ | ~ S = s
		}
		&= 
		\cPP{
			F = A(S) ~ | ~ S = s
		}
		\tagexplain{$F,A(s) \in \cF_s$}
		\nonumber
		\\
		&= 
		\cPP{
			F = f_1 ~ | ~ S = s
		}
		\tagexplain{$A(s) = f_1$}
		\nonumber
		\\
		&= 1/k,
	\end{align}
	and
	\begin{align}
		\cPP{\Loss{\cD_F}{A(S)} \geq 2\alpha ~ | ~ S = s}
		&= 
		\cPP{F \neq A(S) ~ | ~ S = s}
		\nonumber
		\\
		&= 
		\cPP{F \in \{f_2,\dots,f_k\} ~ | ~ S = s}
		\nonumber
		\\
		&= (k-1)/k.
	\end{align}
	Hence, for any $\eta \in \bbR$,
	\begin{align}
		\label{eq:large-estimation-error-for-fixed-s}
		\cPP{
			\left|
				\Loss{\cD_F}{A(S)} - \eta
			\right| \geq \alpha
			~ \big| ~ S = s
		}
		\geq \frac{1}{k}.
	\end{align}
	%Let $\bbD = \{\cD_f: f \in \cF\}$. 
	We conclude that for any estimator $\estimator: \: \left(\cX \times \pmo\right)^m \to \bbR$,
	\begin{flalign*}
		&
		% \PPP{
		% 	\substack{F \sim \uniform{\cF} \\ S \sim (\cD_F)^m}
		% }{
		% 	\bigl|\Loss{\cD_F}{A(S)}
		% 	 - 
		% 	\estimator(S)
		% 	\bigr| \geq \alpha
		% }
		% &
		% \\
		% &
		% \qquad
		% =
		\cPP{
			\left|
				\Loss{\cD_F}{A(S)} - \estimator(S)
			\right| \geq \alpha
		}
		&
		\\
		&
		\qquad 
		\geq
		\sum_{k \in \left\{2,\dots,|\cF|\right\}}
		\cPP{
			\left|
				\Loss{\cD_F}{A(S)} - \estimator(S)
			\right| \geq \alpha
			~ \bigwedge ~ |\cF_S| = k
		}
		&
		\\
		&
		\qquad
		=
		\sum_{k \in \left\{2,\dots,|\cF|\right\}}
		\sum_{s: \:|\cF_s|=k}
		\!\!\!
		\cPP{
			\left|
				\Loss{\cD_F}{A(S)} - \estimator(S)
			\right| \geq \alpha
			~ \big| ~ S = s
		}
		\cdot
		\cPP{S = s}
		&
		\\
		&
		\qquad
		\geq
		\sum_{k \in \left\{2,\dots,|\cF|\right\}}
		\sum_{s: \:|\cF_s|=k}
		\inf_{\eta \in \bbR} \cPP{
			\left|
				\Loss{\cD_F}{A(S)} - \eta
			\right| \geq \alpha
			~ \big| ~ S = s
		} 
		\cdot
		\cPP{S = s}
		&
		\\
		&
		\qquad
		\geq
		\sum_{k \in \left\{2,\dots,|\cF|\right\}}
		\sum_{s: \:|\cF_s|=k}
		\frac{1}{k} \cdot \cPP{S = s}
		\tagexplain{By \cref{eq:large-estimation-error-for-fixed-s}}
		&
		\\
		&
		\qquad
		=
		\sum_{k \in \left\{2,\dots,|\cF|\right\}}\frac{1}{k}\cdot\cPP{|\cF_S| = k}
		&
	\end{flalign*}
	as desired.
\end{proof}

\section{Concentration Bound via Linear Programming}

\begin{lemma}
    \label{lemma:lp-concentration}
	Let $\kmax \in \bbN$, $\vmax \in \bbR$. Let $Z$ be a random variable taking values in $[\kmax]$ such that $\mu = \EE{Z} \in [2, \sqrt{2}+1]$ and $\Var{Z} \leq \vmax$. Then $\PP{Z \in \{2, 3\}}
	\geq 1 - \vmax /2$.
\end{lemma}

We prove this concentration of measure bound using the duality of linear programs (see Section 7.4.1 in \cite{DBLP:books/cu/BV2014} for an exposition of this approach).     

\begin{proof}
	Let $Z' = Z - \mu$. $Z'$ is a random variable with $\EE{Z'} = 0$ and $\Var{Z'} = \Var{Z}$. Furthermore, $\PP{Z \in \{2, 3\}} = \PP{Z' \in \{2 - \mu, 3 - \mu\}}$. We show a lower bound on $\PP{Z' \in \{2 - \mu, 3 - \mu\}}$ across all distribution of $Z'$ with the above moment constraints.

	Indeed, let $X$ be a random variable taking values in $\{1 - \mu, 2 - \mu, \dots, \kmax - \mu\}$ with $\EE{X} = 0$ and $\Var{X} \leq \vmax$ such that $\PP{X \in \{2 - \mu, 3 - \mu\}}$ is minimal. In particular, the distribution of $X$ is a solution to the following minimization problem.
	\begin{flalign*}
		&
		~~
		\min_{\cD_X} \PP{X \in \{2 - \mu,3 - \mu\}}
		&
		\\
		&
		\quad \mathrm{s.t.}
		&
		\\
		&
		\qquad 
		\EE{X} = 0
		&
		\\
		&
		\qquad
		\Var{X} \leq \vmax
		&
	\end{flalign*}
	The minimization problem can be formulated as a linear program with variables $p_k = \PP{X = k - \mu}$ for each $k \in [\kmax]$.
	\begin{flalign*}
		&
		~~
		\min_{\cD_X} p_2 + p_3
		&
		\\
		&
		\quad \mathrm{s.t.}
		&
		\\
		&
		\qquad 
		\sum_{k \in [\kmax]}p_k \geq 1
		&
		\\
		&
		\qquad 
		\sum_{k \in [\kmax]} -p_k \geq -1
		&
		\\
		&
		\qquad 
		\sum_{k \in [\kmax]}p_k\cdot (k-\mu) \geq 0
		&
		\\
		&
		\qquad 
		\sum_{k \in [\kmax]}p_k\cdot (\mu-k) \geq 0
		&
		\\
		&
		\qquad 
		\sum_{k \in [\kmax]}-p_k\cdot (k-\mu)^2 \geq -\vmax
		&
		\\
		&
		\qquad 
		\:\forall k \in [\kmax]: \: p_k \geq 0.
		&
	\end{flalign*}
	This linear program can be represented as
	\begin{flalign*}
		&
		~~
		\min \begin{pmatrix}
			0, 1, 1, 0, \dots, 0 
		\end{pmatrix}
		\cdot p
		&
		\\
		&
		\quad \mathrm{s.t.}
		&
		\\
		&
		\qquad \begin{pmatrix}
			1 & 1 & \dots & 1 \\
			-1 & -1 & \dots & -1 \\
			1-\mu & 2-\mu & \dots & \kmax-\mu \\
			\mu-1 & \mu-2 & \dots & \mu-\kmax \\
			-(1-\mu)^2 & -(2-\mu)^2 & \dots & - (\kmax-\mu)^2 \\
		\end{pmatrix}
		\begin{pmatrix}
			p_1 \\
			\vdots \\
			p_{\kmax} \\
		\end{pmatrix}
		\geq
		\begin{pmatrix}
			1 \\
			-1 \\
			0 \\
			0 \\
			-\vmax
		\end{pmatrix}
		&
		\\
		&
		\qquad ~ p \geq 0.
		&
	\end{flalign*}
	Recall the symmetric duality
	\begin{center}
		\begin{tabular}{l c l}
			$\min c^T x$ & ~ & $\max b^T y$ 
			\\
			s.t. & $~~~~ \leftrightsquigarrow ~~~~ $ & s.t. 
			\\
			\qquad $Ax \geq b$
			&
			~
			&
			\qquad $A^Ty \leq c$
			\\
			\qquad $x \geq 0$
			&
			~
			&
			\qquad $y \geq 0.$
		\end{tabular}
	\end{center}
	Hence, the dual linear program is
	\begin{flalign*}
		&
		~~
		\max \begin{pmatrix}
			1, -1, 0, 0, -\vmax
		\end{pmatrix}
		\cdot y
		&
		\\
		&
		\quad \mathrm{s.t.}
		&
		\\
		&
		\qquad \begin{pmatrix}
			1 & -1 & 1-\mu & \mu-1 & -(1-\mu)^2
			\\
			1 & -1 & 2-\mu & \mu-2 & -(2-\mu)^2
			\\
			1 & -1 & 3-\mu & \mu-3 & -(3-\mu)^2
			\\
			~ & ~ & ~ & $\vdots$ & ~
			\\
			1 & -1 & \kmax-\mu & \mu-\kmax & -(\kmax-\mu)^2
		\end{pmatrix}
		\begin{pmatrix}
			y_1 \\
			\vdots \\
			y_5 \\
		\end{pmatrix}
		\leq
		\begin{pmatrix}
			0 \\
			1 \\
			1 \\
			0 \\
			\vdots \\
			0
		\end{pmatrix}
		&
		\\
		&
		\qquad ~ y \geq 0.
		&
	\end{flalign*}
	A direct calculation shows that the vector
	\[
		y^* = \begin{pmatrix}
			1, 0, \alpha, 0, \frac{1}{2}
		\end{pmatrix},
		\qquad
		\alpha = \frac{1}{\mu-1}
		-
		\frac{\mu-1}{2}
	\]
	is a feasible solution for the dual program for any $\mu \in [2, \sqrt{2}+1]$. 
	The value of the dual program at $y^*$ is $u = 1 - \vmax/2$.
	The weak duality theorem for linear programs implies that $u$ is a lower bound on the value of the primal problem. Hence,
	\[
		\min \PP{X \in \{2 - \mu,3 - \mu\}} \geq u.	
	\]
	This implies that $\PP{Z \in \{2, 3\}} \geq u$, as desired.
\end{proof}

\section{Agreement Between Nearly-Orthogonal Functions}

\begin{claim}
	\label{claim:orthogonal-three-way-agreement}
	Let $\varepsilon > 0$, let $\cX$ be a set, and let $f,g,h: \: \cX \to \pmo$ such that $\{f,g,h\} \in \orthfull{\varepsilon}{\cX}$. Then $\PPP{x \sim \uniform{\cX}}{f(x)=g(x)=h(x)} \leq \frac{1}{4}+\frac{3\varepsilon}{4}$.
\end{claim}

\begin{proof}
	Denote
	\begin{align*}
		a &= \PPP{x \sim \uniform{\cX}}{f(x)=g(x)=h(x)}
		\\
		b &= \PPP{x \sim \uniform{\cX}}{f(x)\neq g(x)=h(x)}
		\\
		c &= \PPP{x \sim \uniform{\cX}}{f(x) = g(x) \neq h(x)}
		\\
		d &= \PPP{x \sim \uniform{\cX}}{f(x) \neq g(x) \neq h(x)}
	\end{align*}
	From $\{f,g,h\} \in \orthfull{\varepsilon}{\cX}$ and \cref{fact:orthogonal-loss},
	\begin{align*}
		a+b &= \PPP{x \sim \uniform{\cX}}{g(x)=h(x)} \leq \frac{1}{2}+\frac{\varepsilon}{2} 
		\\
		a+c &= \PPP{x \sim \uniform{\cX}}{f(x)=g(x)} \leq \frac{1}{2}+\frac{\varepsilon}{2}
		\\
		a+d &= \PPP{x \sim \uniform{\cX}}{f(x)=h(x)} \leq \frac{1}{2}+\frac{\varepsilon}{2}.
	\end{align*}
	Adding these inequalities yields
	\begin{align*}
		3a+b+c+d 
		&\leq
		\frac{3}{2}+\frac{3\varepsilon}{2}.
	\end{align*}
	From the identity $a+b+c+d=1$,
	\begin{align*}
		2a 
		&\leq
		\frac{1}{2}+\frac{3\varepsilon}{2},
	\end{align*}
	so $a \leq \frac{1}{4}+\frac{3\varepsilon}{4}$, as desired.
\end{proof}

\section{Miscellaneous Lemmas}

The following result from \textcite{DBLP:journals/jcss/Achlioptas03} is a variant of a lemma of \textcite{johnson1984extensions}. 

\begin{theorem}[Johnson--Lindenstrauss]
	\label{theorem:johnson-lindenstrauss}
	Let $n,s \in \bbN$, let $\varepsilon,\beta > 0$, and let $V \subseteq \bbR^s$ be a set with cardinality $|V| = n$. Let $d \in \bbN$ such that 
	\[
		d \geq \frac{4 + 2\beta}{\varepsilon^2/2 - \varepsilon^3/3}\ln(n).	
	\]
	Let $R$ be a $d \times s$ random matrix such that each entry is chosen independently and uniformly at random from $\pmo$. Let $f_R: \bbR^s \to \bbR^d$ be given by $f_R(v) = (1/\sqrt{d}) \cdot Rv$. Then
	\[
		\PPPunder{R \sim \uniform{\pmo^{d \times s}}}{\forall u,v \in V: ~ (1-\varepsilon)\|u-v\|_2^2 \leq \|f_R(u)-f_R(v)\|_2^2 \leq (1+\varepsilon)\|u-v\|_2^2} \geq 1-\frac{1}{n^\beta}.
	\]
\end{theorem}

\begin{claim}[Converse to Birthday Paradox]
	\label{claim:birthday-paradox}
	Let $d,m \in \bbN$, and let $\beta \in (0,1)$. If 
	\[
		m \leq \min\left\{\sqrt{d\ln\left(\frac{1}{\beta}\right)},\:\frac{d}{2}\right\}	
	\]
	then $\PPP{X \sim \left(\uniform{[d]}\right)^m}{
			|X| = m
		}	
	\geq \beta$.
\end{claim}

\begin{proof}
	We use the inequality $1-x \geq e^{-x/(1-x)}$, which holds for $x < 1$.
	\begin{align*}
		\PPP{X \sim \left(\uniform{[d]}\right)^m}{
			|X| = m
		} 
		&=
		1\cdot\left(1-\frac{1}{d}\right)\cdot\left(1-\frac{2}{d}\right)\cdots\left(1-\frac{m-1}{d}\right)
		\\
		&\geq 
		\prod_{k=0}^{m-1} \expf{-\frac{k}{d-k}}
		= 
		\expf{- \sum_{k=0}^{m-1} \frac{k}{d-k}}
		\\
		&\stackrel{(*)}{\geq}
		\expf{-\frac{2}{d}\sum_{k=0}^{m-1} k}
		\geq 
		\expf{-\frac{m^2}{d}},
	\end{align*}
	where $(*)$ follows from $m \leq d/2$. Solving $\expf{-\frac{m^2}{d}} \geq \beta$ yields the desired bound. \qedhere
\end{proof}

\begin{theorem}[\cite{hoeffding1963probability}]
	Let $a,b,\mu \in \bbR$ and $m \in \bbN$. Let $Z_1, \dots, Z_m$ be a sequence of i.i.d.\ real-valued random variables and let $Z=\frac{1}{m} \sum_{i=1}^m Z_i$. Assume that $\EE{Z}=\mu$, and for every $i \in [m]$, $\PP{a \leq Z_i \leq b}=1$. Then, for any $\varepsilon>0$,
	\[
		\PP{\left|Z-\mu\right|>\varepsilon} \leq 2 \exp\left(\frac{-2 m \varepsilon^2}{(b-a)^2}\right).
	\]
\end{theorem}

\section{Experiments}

\label{section:experiments}
\subsection{Motivation and Setup}
% Given an algorithm $A$ and a learning task $\mathbb D$, hypothesis stability of $A$ w.r.t. $\mathbb D$ would give rise to the following simple generalization bound: 
% \begin{itemize}
% \item Sample $S\sim \cD^n$, where $\cD \in \mathbb D$.
% \item Train the model, i.e., compute $A(S)$. If the initialization is random, store it.
% \item Compute $A(S')$ (from the same initialization) over a training set $S'$ which is obtained from $S$ by removing $k$ data points uniformly at random without replacement. 
% \item Calculate $L_{S\backslash S'}(A(S'))$, the accuracy over the held out set.
% \end{itemize}
% Now, if the algorithm is $(\alpha,\beta,m,k)$ hypothesis stable, this immediately implies that $\estimator=L_{S\backslash S'}(A(S'))$ is an $(\alpha,\beta,m)$-uniformly tight estimator for $(A,\mathbb D)$.

Here, we examine if there are practical algorithms that admit loss stability or even hypothesis stability with substantial numerical values. To this end, we conduct experiments over a simple neural network architecture across four datasets: MNIST, FashionMNIST, CIFAR10, and CIFAR10 with random labels (figures \ref{fig1}-\ref{fig4}, respectively). Throughout all experiments, we employ one-hidden-layer multi-layer perceptrons with $512$ hidden neurons. We train with SGD across batches of size $1000$ and with momentum $0.9$. For every data set, we train the models across learning rates $0.1$, $0.035$,\footnote{Except for CIFAR10, we present the results only for learning rate 0.1 and 0.01 to prevent clutter. The qualitative results are consistent across all datasets; that is, the curves of learning rate 0.035 lie between the curves of learning rate 0.1 and 0.01.}  and $ 0.01$. We average all the curves over $10$ random seeds (tied for the pairs of networks) and plot the standard deviation for all the curves.

The training procedure is as follows: we train two models in tandem, starting from the same random initialization. The first model is provided with the full training set, whereas the second model has $k=100$ data points removed from its training set. These points are drawn uniformly at random before the beginning of the training, and fixed thereafter. After each epoch, we evaluate the training accuracy, test accuracy and hypothesis stability, i.e., the agreement between the two models (which we calculate across the test set).

We set our main focus on the agreement of the models since the most amenable way to show loss stability might be by way of proving hypothesis stability. The latter can perhaps be mathematically proven in the case of neural networks by analyzing the stability of the training dynamics under two slightly different training sets.

\subsection{Results}
Across all experiments, the training and test accuracy of the model pairs are essentially identical throughout the training process. This suggests that at least simple models are loss stable across vision tasks.
In order to reduce visual clutter, we hence only plot training and test accuracy of the first model (which has access to the full training set), respectively.

We observe higher agreement for simpler data sets and smaller learning rates. For example, the learning rate has a considerable effect on agreement for CIFAR10 ($\approx 0.65$ for learning rate $0.1$ vs $\approx 0.8$ for learning rate $0.01$).

\begin{figure}[!ht]
    \centering
    \includegraphics[width=0.95\textwidth]{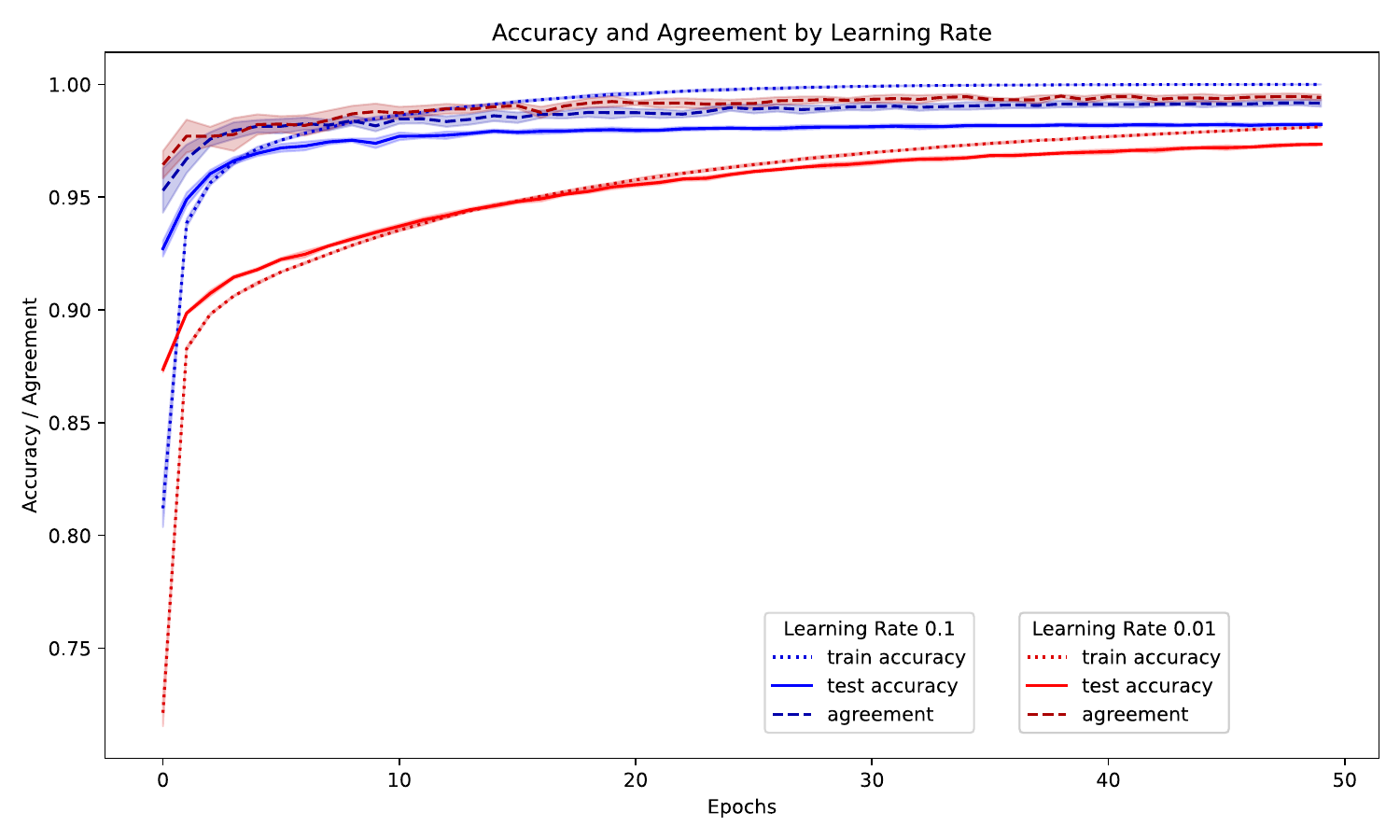}
    \caption{MNIST}\label{fig1}
\end{figure}

\begin{figure}[!ht]
    \centering
   \includegraphics[width=0.95\textwidth]{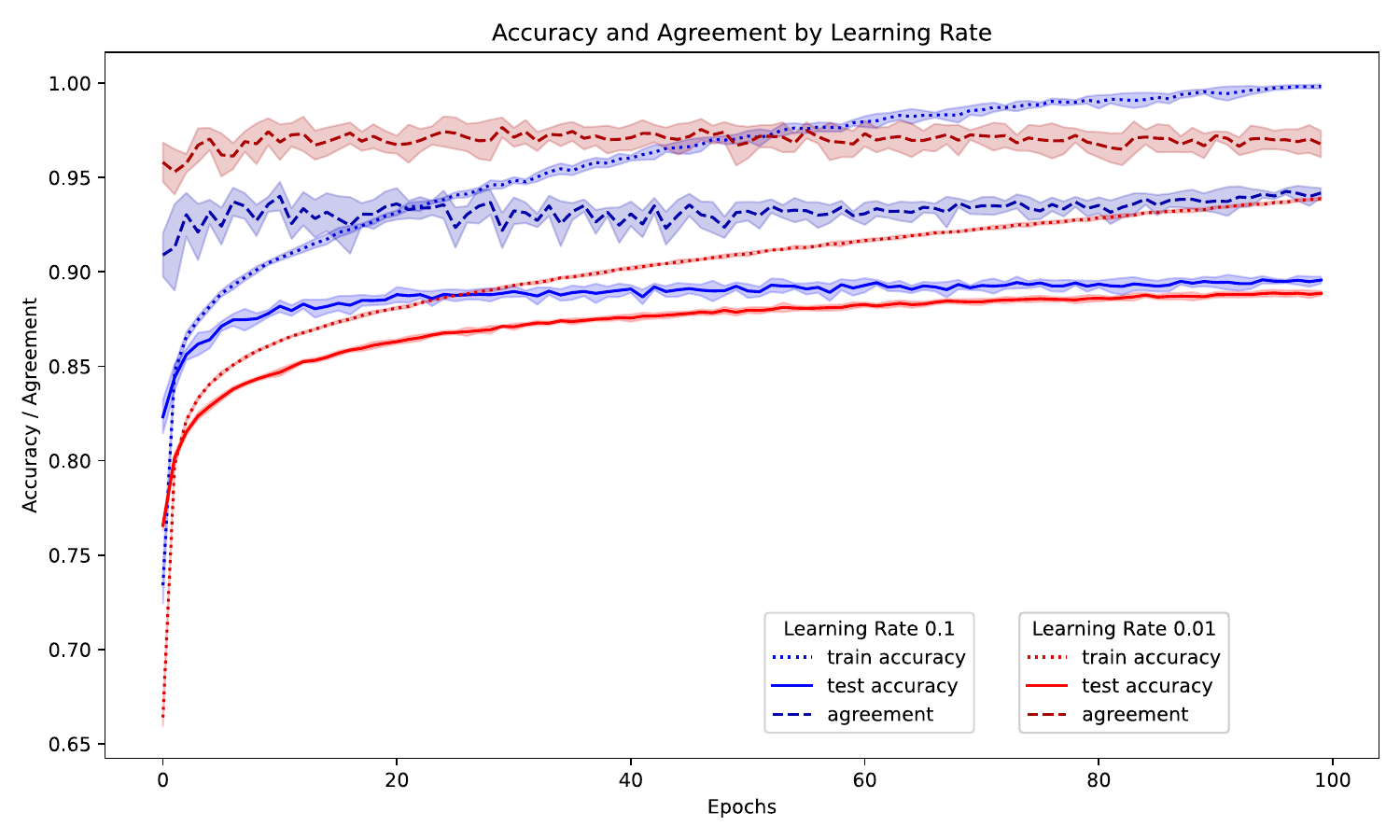}
    \caption{FashionMNIST}\label{fig2}
\end{figure}

\begin{figure}[!ht]
    \centering
    \includegraphics[width=0.95\textwidth]{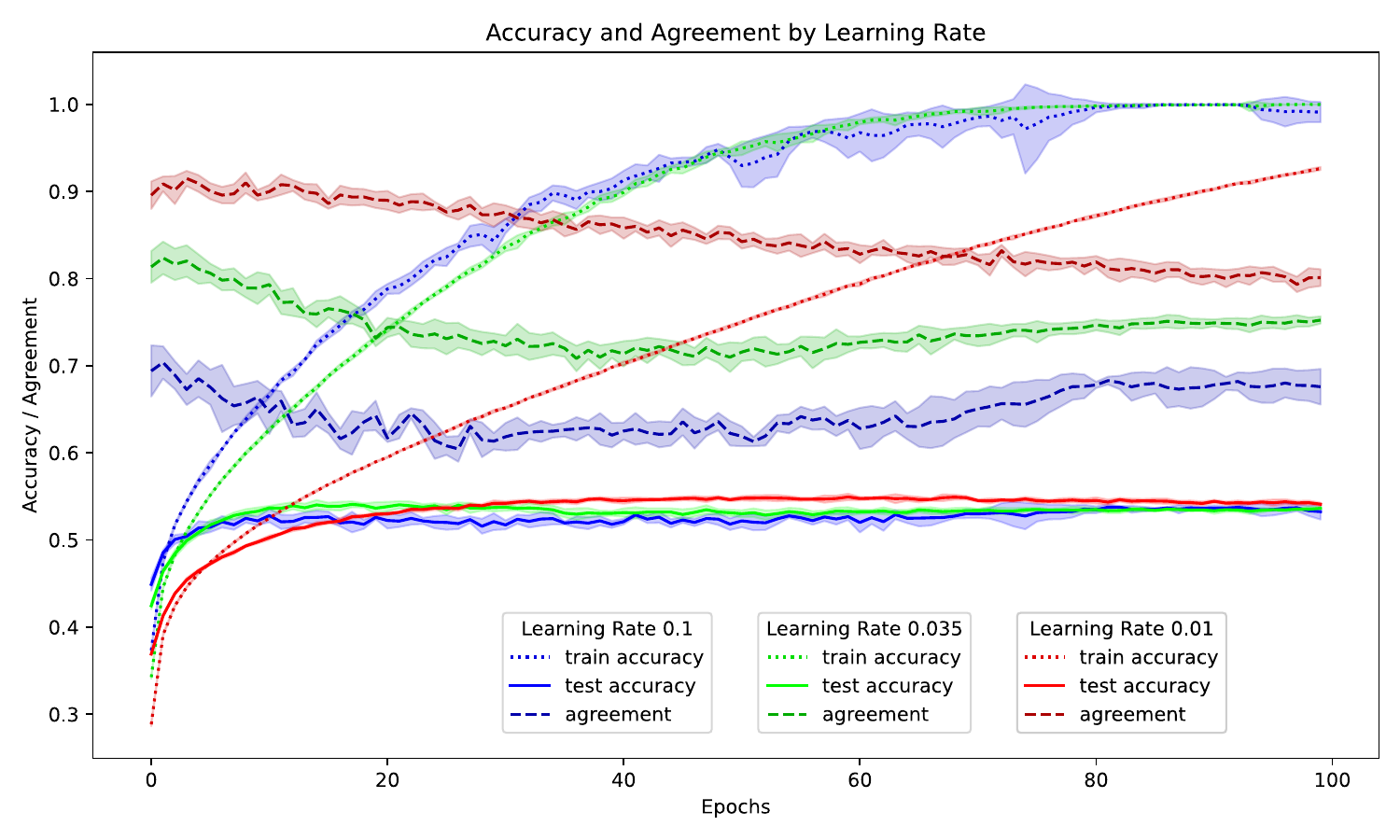}
    \caption{CIFAR10}\label{fig3}
\end{figure}

\begin{figure}[!ht]
    \centering
    \includegraphics[width=0.95\textwidth]{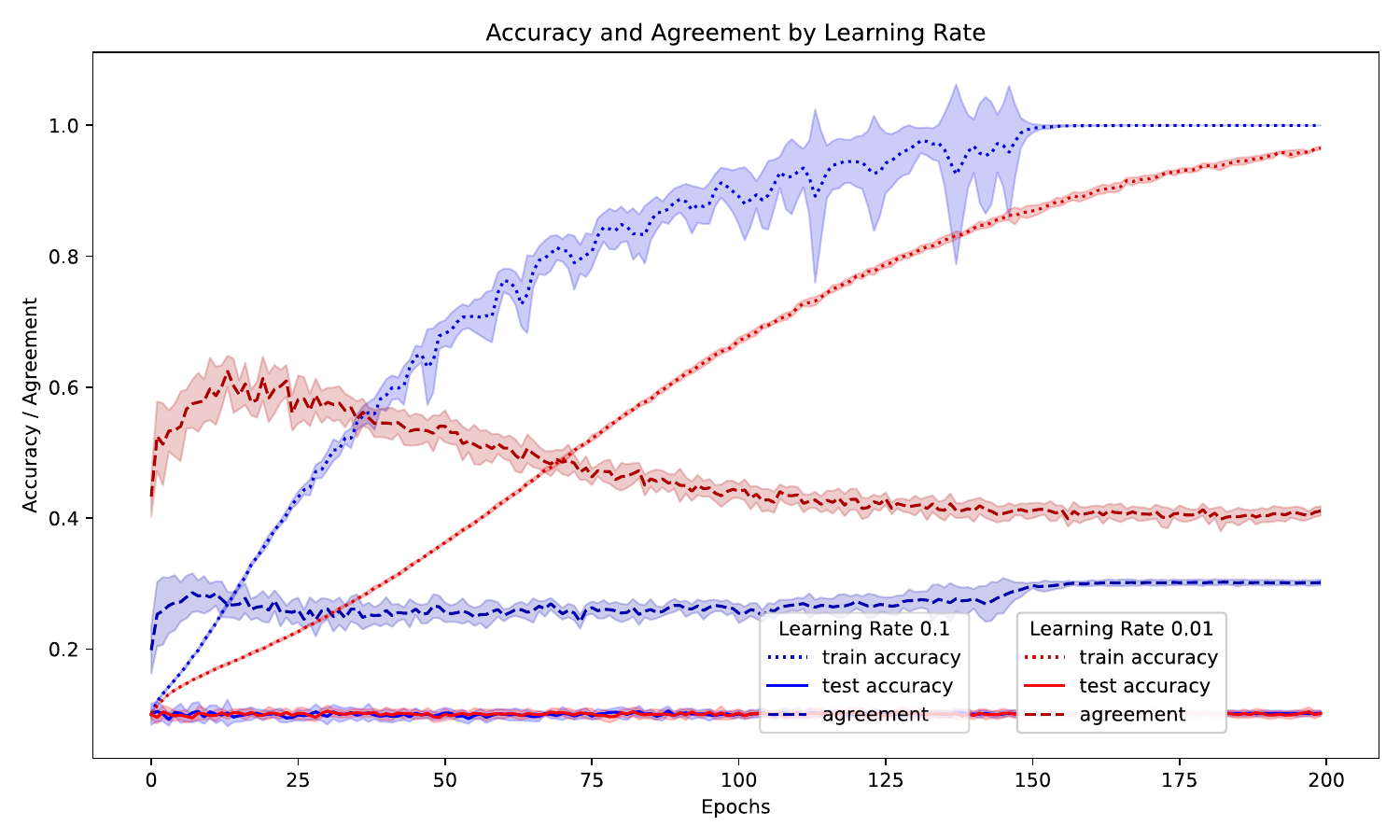}
    \caption{CIFAR10 with random labels}\label{fig4}
\end{figure}

The key takeaway from Figures \ref{fig1} through \ref{fig4} is that the agreement is consistently higher than the test accuracy. This relationship ensures that when applying the estimation procedure outlined in Theorem~\ref{theorem:stablity-implies-estimability}, we can avoid vacuous predictions of perfect accuracy. In the scenarios presented, the estimated accuracy will always be bounded away from 1, as it can be expressed as \textit{test error + (1 - agreement)}. For instance, with a learning rate of 0.01, the maximum estimated accuracies are: $98\%$ for MNIST (compared to $97.5\%$ test accuracy), $90\%$ for FashionMNIST ($87\%$ test accuracy), $72\%$ for CIFAR10 ($52\%$ test accuracy), and $65\%$ for CIFAR10 with random labels ($10\%$ test accuracy). These results illustrate a strong correlation between stability estimation and data complexity.

% The main takeaway from figures \ref{fig1}-\label{fig4} is that the agreement is always higher than the test accuracy. This guarantees that when using the estimation procedure in Theorem~\ref{theorem:stablity-implies-estimability}, the estimation will never give false hopes of perfect accuracy; the estimation of accuracy in the scenarios we presented will always be bounded away from 1, since the estimation can be at most \textit{test error + (1-agreement)}. Specifically, for learning rate 0.01, the estimation of the accuracy can be at most: $98\%$ (vs. $97.5\%$ test accuracy) for MNIST, $90\%$ (vs. $87\%$ test accuracy) for FashionMNIST, $72\%$ (vs. $52\%$ test accuracy) for CIFAR10, and $65\%$ (vs. $10\%$ test accuracy) for CIFAR10 with random labels. We see that the stability estimation correlate well with the data complexity.
% Generally, the value of agreement plateaus faster than both the training and test accuracy. On the other hand, in some settings, agreement admits larger small scale fluctuations around its mean value. For example, for FashionMNIST with learning rate $0.01$, we reach large agreement (around $0.97$) already after $5$ epochs, but we observe continued small scale fluctuations around this value throughout training.
\newpage

\end{document}